%% file: main.tex
\documentclass[10pt,twocolumn,letterpaper]{article}

\usepackage[pagenumbers]{cvpr} % To force page numbers, e.g. for an arXiv version


\definecolor{cvprblue}{rgb}{0.21,0.49,0.74}
\usepackage[pagebackref,breaklinks,colorlinks,citecolor=cvprblue]{hyperref}
\usepackage{amsmath,amssymb,amsfonts}
\usepackage{algorithmic}
\usepackage{algorithm}
\usepackage{graphicx}
\usepackage{multirow}
\usepackage{marvosym}
\usepackage{xcolor}
\renewcommand{\thefootnote}{\fnsymbol{footnote}}
\usepackage{nicematrix}
\usepackage{float}
\usepackage[hang,flushmargin]{footmisc}
\definecolor{lightgreen}{RGB}{231,245,234}
\definecolor{lightpurple}{RGB}{236,223,247}
\definecolor{lightred}{RGB}{255,204,153}
\definecolor{tabhighlight}{HTML}{e5e5e5}
\usepackage{bm}
\usepackage{colortbl}
\usepackage{booktabs}
\usepackage{pifont}
\newcommand{\parhead}[1]{\noindent\textbf{#1}\,}

\newcommand{\cmark}{\ding{51}} % Checkmark symbol
\newcommand{\xmark}{\ding{55}} % Crossmark symbol

\newcommand*{\affaddr}[1]{#1} % No op here. Customize it for different styles.

\newcommand{\tablestyle}[2]{\setlength{\tabcolsep}{#1}\renewcommand{\arraystretch}{#2}\centering\footnotesize}
\def\BibTeX{{\rm B\kern-.05em{\sc i\kern-.025em b}\kern-.08em
    T\kern-.1667em\lower.7ex\hbox{E}\kern-.125emX}}
\def\paperID{10145} % *** Enter the Paper ID here
\def\confName{CVPR}
\def\confYear{2026}

\title{MedCLIPSeg: Probabilistic Vision-Language Adaptation for Data-Efficient and Generalizable Medical Image Segmentation}
\author{%
    \makebox[\linewidth][c]{%
        Taha Koleilat\textsuperscript{\Letter}
        \hspace{3em} Hojat Asgariandehkordi
        \hspace{3em} Omid Nejati Manzari
    }\\[0.5ex]
    \makebox[\linewidth][c]{%
        Berardino Barile
        \hspace{3em} Yiming Xiao\textsuperscript{$\dagger$}
        \hspace{3em} Hassan Rivaz\textsuperscript{$\dagger$}
    }\\[1ex]
    \affaddr{Concordia University, Montreal, Canada}
    \\[1ex]
    \url{https://tahakoleilat.github.io/MedCLIPSeg}
}

\begin{document}
\maketitle
\input{sec/0_abstract}  
\renewcommand{\thefootnote}{}
\footnote{\Letter\ Corresponding Author: \href{mailto:taha.koleilat@mail.concordia.ca}{taha.koleilat@mail.concordia.ca}}
\footnote{$\dagger$ Co-senior authors}
\input{sec/1_intro}
\input{sec/2_method}
\input{sec/3_results}
\input{sec/4_conclusion}
{
    \small
    \bibliographystyle{ieeenat_fullname}
    \bibliography{main}
}

\input{sec/X_suppl}

\end{document}

%% file: sec/0_abstract.tex
\begin{abstract}
Medical image segmentation remains challenging due to limited annotations for training, ambiguous anatomical features, and domain shifts. While vision-language models such as CLIP offer strong cross-modal representations, their potential for dense, text-guided medical image segmentation remains underexplored. We present \texttt{MedCLIPSeg}, a novel framework that adapts CLIP for robust, data-efficient, and uncertainty-aware medical image segmentation. Our approach leverages patch-level CLIP embeddings through probabilistic cross-modal attention, enabling bidirectional interaction between image and text tokens and explicit modeling of predictive uncertainty. Together with a soft patch-level contrastive loss that encourages more nuanced semantic learning across diverse textual prompts, \texttt{MedCLIPSeg} effectively improves data efficiency and domain generalizability. Extensive experiments across 16 datasets spanning five imaging modalities and six organs demonstrate that \texttt{MedCLIPSeg} outperforms prior methods in accuracy, efficiency, and robustness, while providing interpretable uncertainty maps that highlight local reliability of segmentation results. This work demonstrates the potential of probabilistic vision-language modeling for text-driven medical image segmentation.
\end{abstract}

%% file: sec/1_intro.tex
\section{Introduction}
\label{sec:intro}

Accurate and trustworthy medical image segmentation remains a cornerstone for diagnosis, treatment planning, and quantitative clinical follow-up. Yet progress is often constrained by three persistent obstacles. First, \textit{expert annotations} of segmentation ground truths are expensive and often inconsistent across raters, restricting the quality of supervised learning. Second, lesions and organs can exhibit \textit{ambiguous boundaries} due to gradual intensity transitions or partial-volume effects that make clear decision-making difficult. Third, common \textit{domain shifts} in scans due to variations in scanners, acquisition protocols, and patient populations cause models trained on limited in-distribution (ID) data to fail when exposed to out-of-distribution (OOD) conditions. These issues expose an urgent need for segmentation systems that are simultaneously data-efficient, uncertainty-aware, and generalizable across domains.

\begin{figure}
    \centering
    \includegraphics[width=0.98\linewidth]{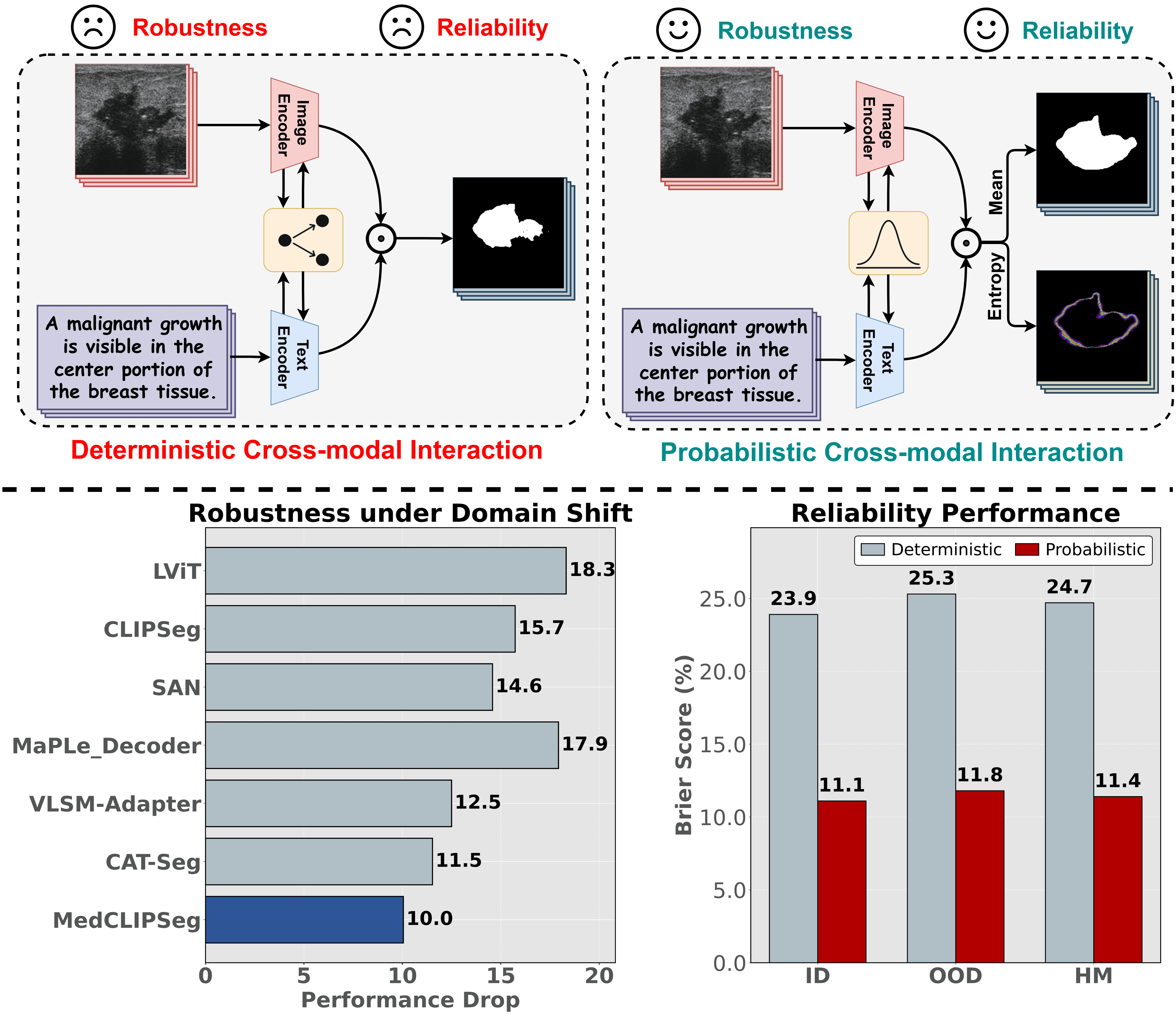}
    \caption{{\color{blue}\textbf{\textit{(Top)}}}: \textbf{Comparison between deterministic and probabilistic cross-modal fusion techniques in CLIP adaptation for text-driven segmentation.} Probabilistic formulation models variability in visual–textual representations as distributions, enabling more robust feature alignment. {\color{blue}\textbf{\textit{(Bottom)}}}: \textbf{Robustness} and \textbf{Reliability} plots over ID and OOD data show improved generalization, with smaller out-of-domain performance drops and better calibration of predicted confidence, reflected by lower Brier scores.
    }
    \label{fig:figure1}
\end{figure}

\noindent
In medical image segmentation, U-Net and its variants \cite{ronneberger2015unet, isensee2021nnu} have driven major advances, by exploiting convolutional neural network (CNN)'s inductive biases for efficient feature learning and/or Vision Transformer (ViT) architectures \cite{chen2021transunet, cao2022swin, hatamizadeh2022unetr} for long-range dependencies. Despite their success, these models depend on extensive pixel-wise supervision and most operate deterministically. Prior work shows that such segmentation networks, including U-Nets, are systematically \emph{over-confident}, particularly for out-of-distribution inputs and those with fuzzy tissue boundaries \cite{ao2023two,ye2023mitigating}, yielding unreliable segmentation results without warning mechanisms. This highlights the issues in conventional feature learning with deterministic approaches, which fail to properly account for ambiguity or local disagreement among features. Learning methods that can adaptively weigh evidence and/or modulate attention based on contextual reliability of data distribution would be hopeful to address this \cite{mehrtash2020confidence}. Additionally, alternative methods that move beyond traditional fully supervised approaches to mitigate the heavy reliance on expert training data while enhancing generalizability and interactability are highly desirable. 

By aligning images and texts through large-scale contrastive pretraining, vision–language models (VLMs) such as CLIP \cite{radford2021learning} and its biomedical variants \cite{biomedclip, khattak2024unimed} offer a promising paradigm towards more label-efficient and generalizable medical image segmentation with the potential for intuitive natural language-driven user-interaction. Notably, recent studies show that CLIP’s patch tokens can encode spatial semantics \cite{zhou2022extract, gandelsman2023interpreting}, suggesting its capacity for dense localization even without pixel-level supervision. Yet, in medical domain, where more subtle visual differences and fine-grained descriptions limit multi-modal alignment \cite{zhao2025clip}, the dense localization capability in VLMs remains weaker \cite{farooq2025localization}, but can be potentially strengthened through \emph{more nuanced image–text mapping} \cite{koleilat2025medclip} and \emph{tailored, task-specific representation adaptation} without degrading generalization \cite{huang2024adapting}. Practically, clinical descriptions are far easier to obtain than pixel-level masks, making VLMs appealing in low-data regimes, where textual supervision compensates for limited annotations \cite{koleilat2025biomedcoop}. While most methods emphasize prompt learning, decoder tuning, or unidirectional text-to-vision modulation \cite{poudel2023exploring, yang2022lavt, li2023lvit}, \emph{deep cross-modal fusion} was shown to better support CLIP adaptation and spatial grounding \cite{zhang2024evf, khattak2023maple}. Yet, CLIP models with deterministic representations still remain over-confident on OOD data \cite{Murugesan2024}, motivating probabilistic CLIP formulation \cite{cheng2025vamp}, which remains underexplored for medical image segmentation.

In this context, a key question emerges: \textit{how can we formulate cross-modal attention to be uncertainty-aware for CLIP-based segmentation?} This is particularly relevant in practical medical AI adoption, where model credence and transparency are crucial, while over-confident deterministic AI models fail to meet the needs. Building on existing insights, we propose \texttt{MedCLIPSeg}, a text-driven medical image segmentation framework that adapts CLIP with probabilistic, bidirectional vision–language representation fusion. Its core component, the \emph{Probabilistic Vision–Language (\texttt{PVL}) adapter} learns confidence-weighted attention between image patches and text tokens within CLIP’s multiple deep encoding layers. Confidence-aware attention scores based on variational modeling of \textit{Keys} reduce over-confidence, and Monte Carlo sampling of \textit{Value} distributions yields both mean segmentation masks and associated pixel-level uncertainty maps for user interpretation. Specifically, our probabilistic modeling of the \textit{Keys} and \textit{Values} in cross-modal attention naturally captures \textit{aleatoric uncertainty} from ambiguous image features and \textit{epistemic uncertainty} arising from unseen domains, leading to better accuracy, improved calibration, and enhanced robustness, in line with \cite{guo2017calibration, kendall2017uncertainties}. As illustrated in Fig.~\ref{fig:figure1}, in contrast to \texttt{MedCLIPSeg}'s probabilistic adaptation, the deterministic variant leads to inaccurate segmentation and over-confidence that manifests as poor calibration and significant performance drops under domain shift. Furthermore, to maintain data efficiency, we preserve CLIP’s pretrained encoders and introduce a soft patch-level contrastive loss that refines image–text alignment for dense prediction under limited supervision. In summary, our main contributions include:

\begin{enumerate} 
\item \textbf{Bidirectional representation-level fusion} that enhances data-efficiency and robustness through novel vision-language interaction adapters while preserving CLIP’s parameters, guided by a soft contrastive loss.

\item \textbf{Probabilistic cross-modal attention} with variational \textit{Key} and \textit{Value} formulation to enable uncertainty-aware learning to improve accuracy and generalizability.

\item \textbf{Pixel-level uncertainty maps} by sampling \textit{Values} from learnt probability distributions in VLM attentions to offer intuitive reliability visualization for clinical review. 

\item \textbf{Comprehensive evaluation} against SOTA methods for medical image segmentation on five modalities and six organs, assessing data efficiency, domain generalizability, and model sub-component performance to provide insights into the proposed framework.

\end{enumerate}

%-------------------------------------------------------------------------
\section{Related Work}
\label{sec:related-work}
\parhead{Medical Image Segmentation:} Medical image segmentation has traditionally relied on vision-only architectures. CNNs established the foundation, with U-Net \cite{ronneberger2015unet} introducing the long skip connections, inspiring UNet++ \cite{zhou2018unet++}, Attention U-Net \cite{oktay2018attention}, and nnUNet \cite{isensee2021nnu}. Other variants, including the DeepLab series \cite{chen2017rethinking}, improved multi-scale context modeling through dilated convolutions and pyramid pooling. The advent of ViTs further advanced segmentation by incorporating long-range dependencies. TransUNet \cite{chen2021transunet} fused CNN backbones with Transformer encoders, while Swin-UNet \cite{cao2022swin} improved efficiency using shifted windows. Building on these designs, hybrid models such as HiFormer \cite{heidari2023hiformer} and UNETR \cite{hatamizadeh2022unetr} demonstrated strong performance across multiple benchmarks. Despite these advances, vision-only approaches often rely heavily on low-level appearance features and show limited robustness to domain shifts across scanners and imaging protocols. This motivates the use of high-level semantic priors, particularly via cross-modal learning to improve generalizability.

\medskip
\parhead{Vision-Language Models:} Vision-language models have gained interest in biomedical domains. As CLIP \cite{radford2021learning} and ALIGN \cite{jia2021scaling} demonstrated strong zero-shot transfer across different visual tasks, they inspired biomedical variants such as BiomedCLIP \cite{biomedclip}, PubMedCLIP \cite{eslami2021doesclipbenefitvisual}, and UniMedCLIP \cite{khattak2024unimed}, which leverage clinical image-text corpora for domain-specific learning. While these models provide robust global alignment, they often require further adaptation to capture the fine-grained semantics of anatomy and pathology. Typical parameter-efficient adaptation techniques include prompt tuning (e.g., CoOp \cite{zhou2022learning}, CoCoOp \cite{zhou2022conditional}, and MaPLe \cite{khattak2023maple}) and low-rank-based model updates (e.g., CLIP-LoRA \cite{zanella2024low} and CLIP-SVD \cite{koleilat2025singular}). Tailored for biomedical visions, methods such as DCPL \cite{cao2024domain} and BiomedCoOp \cite{koleilat2025biomedcoop} incorporate domain priors and knowledge distillation to enhance adaptation under limited supervision. More recently, probabilistic fine-tuning frameworks such as CLAP4CLIP \cite{jha2024clap4clip} and ProbVLM \cite{upadhyay2023probvlm} incorporate embedding uncertainty to better manage the many-to-one mappings between vision and language, improving calibration and cross-domain generalization. Despite these advances, most biomedical VLMs focus on classification or retrieval, with limited exploration of spatially grounded tasks, such as segmentation, which is more challenging.

\medskip
\parhead{Prompt-based Segmentation:}
For natural images, only a few methods extend VLMs to image segmentation. CLIPSeg \cite{luddecke2022image} and CRIS \cite{wang2022cris} append lightweight decoders to frozen CLIP encoders, while LAVT \cite{yang2022lavt} fuses textual embeddings into Transformer layers through cross-attention, modulating visual features throughout the encoder. Further approaches such as DenseCLIP~\cite{rao2022denseclip}, ZegCLIP~\cite{zhou2023zegclip}, and SAN~\cite{xu2023side} enhance fine-grained localisation, while the recent CAT-Seg~\cite{cho2024cat} stands out with strong state-of-the-art performance in open-vocabulary segmentation. On the other hand, the Segment Anything Model (SAM) \cite{kirillov2023segment} introduced a promptable foundation for general-purpose segmentation but lacks explicit natural language interaction and conditioning, and remains limited to drawing-based prompts. Although effective for natural images, these methods struggle in biomedical contexts, where images lack contextual diversity, exhibit high inter-class similarity, and feature ambiguous boundaries. On the other hand, medical adaptations, like LViT \cite{li2023lvit} and Ariadne’s Thread \cite{zhong2023ariadne}, integrated BERT textual embeddings into ViT architectures. MedSAM \cite{medsam} and recent prompt-based work \cite{koleilat2024medclip, koleilat2025medclip, rasaee2025groundingdino, wong2024scribbleprompt, spiegler2025textsam} still rely on geometric prompts at different intermediate stages, introducing potential instability. More recently, few-shot medical image segmentation frameworks, such as UniverSeg \cite{butoi2023universeguniversalmedicalimage}, MultiverSeg \cite{wong2025multiverseg}, and Iris \cite{gao2025show}, leverage a small support set of image–label pairs to segment unseen classes and modalities without additional training. BiomedParse \cite{biomedparse} further explores structured knowledge parsing across modalities through the use of natural language. However, methods for adapting VLMs such as CLIP for dense biomedical prediction remain limited. Poudel \textit{et al.} \cite{poudel2023exploring} applied CLIPSeg and CRIS with frozen CLIP encoders and new decoders, but domain-specific models such as BiomedCLIP \cite{biomedclip} offered no gains, highlighting the weakness of naïve adaptations. VLSM-Adapter \cite{dhakal2024vlsm} offers parameter-efficient VLM adaptation, and CausalCLIPSeg \cite{chen2024causalclipseg} introduces multi-modal causal adaptations, but evaluations on multiple datasets are not provided, and we show its weakness in domain generalization (Table \ref{tab:domain_generalization}). Unlike existing approaches, our method preserves CLIP’s pretrained parameters while introducing probabilistic, bidirectional vision-language fusion to enhance robustness and clinical reliability in biomedical segmentation.

%% file: sec/2_method.tex
\section{Methodology}
\label{sec:methods}
\begin{figure*}
    \centering
    \includegraphics[width=0.98\linewidth]{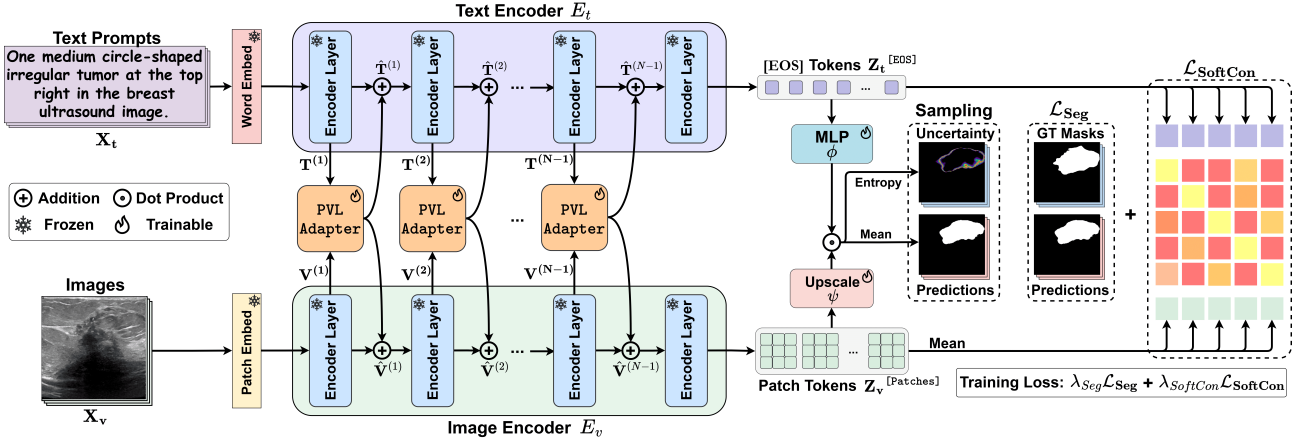}
    \caption{
        Overview of the proposed \texttt{MedCLIPSeg} framework for text-driven medical image segmentation. 
        The model extends CLIP with {\setlength{\fboxsep}{0.7pt}\colorbox{lightgreen}{vision}} and 
        {\setlength{\fboxsep}{0.7pt}\colorbox{lightpurple}{language}} encoders connected via 
        {\setlength{\fboxsep}{0.7pt}\colorbox{lightred}{\texttt{PVL Adapters}}}, which perform confidence-weighted image–text fusion at multiple deep layers. Segmentation and uncertainty maps arise from the \emph{mean} and \emph{entropy} of posterior samples, with a soft patch-level contrastive loss.}
    \label{fig:framework}
\end{figure*}

\noindent Our \texttt{MedCLIPSeg} framework is presented in Fig. \ref{fig:framework}.

\begin{figure}
    \centering
    \includegraphics[width=0.98\linewidth, height=0.45\linewidth]{figures/MedCLIPSeg_PVL.png}
    \caption{
        Illustrations of \textbf{\texttt{PVL Adapter}} and \textbf{ $\texttt{Attn}_{\texttt{PVL}}$.}}
    \label{fig:PVL_adapter}
\end{figure}

\subsection{CLIP Overview}
Our text-driven segmentation framework builds upon the CLIP architecture \cite{radford2021learning}, which comprises two Transformer-based encoders: a vision encoder $\bm{E_v}$ and a text encoder $\bm{E_t}$. They encode images and text into a shared $D$-dimensional space for cross-modal alignment. For a batch of $B$ RGB images, the input is represented as $\bm{X_v} \in \mathbb{R}^{B \times 3 \times H \times W}$. The vision encoder first partitions each image into $P$ non-overlapping patches, projects each to a $D$-dimensional embedding, and prepends a learnable \texttt{[CLS]} token. This results in a sequence of $(P+1)$ tokens:
\begin{equation}
\bm{Z_v} = \bm{E_v}(\bm{X_v}) \in \mathbb{R}^{B \times (P+1) \times D}
\end{equation}
The \texttt{[CLS]} token output, $\bm{Z_v}^{\texttt{[CLS]}} \in \mathbb{R}^{B \times D}$, serves as the global image representation in the shared CLIP space. Notably, prior works \cite{zhou2022extract, gandelsman2023interpreting} have shown that, although CLIP’s objective aligns only the \texttt{[CLS]} tokens, the contrastive vision-language pre-training implicitly shapes the patch token representations to encode semantically meaningful spatial features with emerging textual correlation.

\noindent For the textual branch, tokenized prompts $\bm{X_t} \in \mathbb{R}^{B \times L}$ of length $L$ are embedded into $D$-dimensional vectors and processed by the text encoder. Unlike the vision encoder’s use of a \texttt{[CLS]} token, CLIP takes the output at the \texttt{[EOS]} position as the global text representation:
\begin{equation}
\bm{Z_t} = \bm{E_t}(\bm{X_t}) \in \mathbb{R}^{B \times L \times D}
\end{equation}
Here, the \texttt{[EOS]} output $\bm{Z_t}^{\texttt{[EOS]}} \in \mathbb{R}^{B \times D}$ acts as the compact textual embedding in the joint space. The global embeddings enable cross-modal similarity, while the patch tokens from $\bm{Z_v}$ preserve fine-grained spatial information. In our framework, the global text embedding $\bm{Z_t}^{\texttt{[EOS]}}$ is used as a query and compared via dot product with each vision patch token, producing segmentation logits that guide segmentation according to the natural language prompt.

\subsection{Probabilistic Multi-modal Adaptation}
To enable efficient and confidence-aware multimodal fusion-based CLIP adaptation for biomedical segmentation, we propose the \emph{Probabilistic Vision-Language Adapter} (\texttt{PVL Adapter}), a novel probabilistic framework that bridges CLIP’s vision and language encoders at the representation level. Each \texttt{PVL} module performs bidirectional, probabilistic interaction between image and text tokens. The architecture of \texttt{PVL Adapter} is depicted in Fig.~\ref{fig:PVL_adapter}. 

\medskip
\parhead{Downward Projection:} Given visual tokens $\mathbf{V}^{(n)} \in \mathbb{R}^{B \times T_v \times D_v}$ and text tokens $\mathbf{T}^{(n)} \in \mathbb{R}^{B \times T_t \times D_t}$ at Layer $n$ with total layers amounting to $N$, the \texttt{PVL} adapter first projects both modalities to a shared lower-dimensional space $D_s$ with $W^{\downarrow^{(n)}}_v$ $\in$ $\mathbb{R}^{D_v \times D_s}$ and $W^{\downarrow^{(n)}}_t$ $\in$ $\mathbb{R}^{D_t \times D_s}$:  
\begin{equation}
\mathbf{v}^{(n)} = \mathbf{V}^{(n)} W^{\downarrow^{(n)}}_v, \quad \mathbf{t}^{(n)} = \mathbf{T}^{(n)} W^{\downarrow^{(n)}}_t
\end{equation}

\medskip
\parhead{QKV parameterization:} Inspired by \cite{kendall2017uncertainties}, we extend the standard attention formulation to incorporate uncertainty in both \textit{Keys} and \textit{Values} by modeling them as probability distributions with learnable means and variances. These variances represent data ambiguity, allowing the model to encode inherent noise in the input representations. This probabilistic design enables the attention module to downweight uncertain tokens and sample value representations for stochastic modeling. This attention module $\texttt{Attn}_{\texttt{PVL}}$ takes as input a query sequence $X$ and a context sequence $Z$ and outputs a fused output $Y$ as $Y=\texttt{Attn}_{\texttt{PVL}}(X,Z)$. The input query and context sequences are transformed using the Query (Q), Key (K), Value (V), and Output (O) projection matrices as follows:
\begin{equation}
Q = X \cdot W_Q
\end{equation}
\noindent
where $X \in \mathbb{R}^{B \times T_q \times D_s}$ is the input query sequence, $W_Q \in \mathbb{R}^{D_s \times D_a}$ is the learnable projection matrix, and $Q$ is the transformed query to compute the attention score.

\noindent Keys and Values are projected into both a mean and a log-variance representation:
\begin{equation}
[K_\mu,\, K_{\log\sigma^2}] = Z \cdot W_K, \quad [V_\mu,\, V_{\log\sigma^2}] = Z \cdot W_V
\end{equation}
\noindent
where $Z \in \mathbb{R}^{B \times T_k \times D_s}$ is the context sequence and $W_K \in \mathbb{R}^{D_s \times 2D_a}$ produces two $D_a$-dimensional splits for mean and log-variances.

\noindent To convert the predicted log-variances into original variance values, we avoid the numerically unstable $\exp(\cdot)$ and instead use the \texttt{softplus} $\zeta(.)$ activation, which is smoother and less prone to instabilities:
\begin{equation}
K_{\sigma}^2 = \zeta(K_{\log\sigma^2}), \quad V_{\sigma}^2 = \zeta(V_{\log\sigma^2})
\end{equation}
\noindent Here, $K_{\sigma}^2$ and $V_{\sigma}^2$ represent the variance terms for the \textit{Key} and \textit{Value} distributions.

\medskip
\parhead{Confidence-weighted attention:}
Our proposed attention score considers two terms: a mean similarity $S_{\mu}$ and a variance-based confidence penalty $S_{\sigma}^2$. 
Each score \(S_{ij}\) corresponds to the dot product between a deterministic query vector \(Q_i\) 
and a probabilistic key \(K_j \sim \mathcal{N}(K_{\mu,j},K_{\sigma,j}^2 )\), assuming feature-wise independence. This yields a probabilistic attention score with mean and variance given by:
\begin{equation}
S_{\mu} = \frac{Q K_\mu^\top}{\sqrt{D_a}}, \quad 
S_{\sigma}^2 = \frac{Q^{\circ 2}(K_{\sigma}^2)^\top}{D_a},
\end{equation}
\noindent where \(Q^{\circ 2}\) denotes the element-wise square of \(Q\). The variance term quantifies how key uncertainty interacts with the magnitude of query features, 
allowing uncertain tokens to be adaptively downweighted. The final attention weights are computed as:
\begin{equation}
\label{eq:difference}
A = \{A_{ij}\} = \texttt{softmax}\bigl(S_{\mu} - \beta S_{\sigma}\bigr)\:with 
\end{equation}
\begin{equation}
\label{eq:weighting}
A_{ij}
= \frac{\texttt{exp}~\!\big(S_{\mu,ij}\big) / \omega_{ij}}
{\sum_{r} \texttt{exp}~\!\big(S_{\mu,ir}\big) / \omega_{ir}},
\quad \omega_{ij} = \texttt{exp}~\!\big(\beta\,S_{\sigma,ij}\big)
\end{equation}

\noindent where $\beta = 2.35$ (indicates the full width at half maximum of a Gaussian distribution) scales the confidence penalty, down-weighting overconfident attention responses. This dynamic probabilistic weighting allows the model to emphasize reliable evidence and downweight uncertain signals, thereby regularizing feature learning and improving \emph{out-of-distribution generalization}, which is crucial for medical data with large domain variability. Conceptually, Eq.~\ref{eq:difference} can be interpreted as a \emph{variance-aware extension} of the standard attention mechanism, where attention scores are weighted by an input-dependent uncertainty term (Eq.~\ref{eq:weighting}); when $\beta = 0$, the formulation naturally reduces to the conventional deterministic attention.

\medskip
\parhead{Value sampling:} 
To model uncertainty in the \textit{Value} representations, we draw samples from their learned probability distribution \(\mathcal{N}(V_{\mu},V_{\sigma}^2 )\). Each sample is obtained via the reparameterization trick:
\begin{equation}
V_{\mathrm{sample}} = V_\mu + \epsilon \odot V_{\sigma}, 
\quad \epsilon \sim \mathcal{N}(0, I).
\end{equation}
\noindent where $\odot$ denotes the Hadamard product.

\medskip

\noindent During training, we adopt a stochastic regime by sampling only once to compute the attended output as:
\begin{equation}
O = A \cdot V_{\mathrm{sample}}.
\end{equation}

\noindent At test time, we perform multiple stochastic forward passes to sample from the model’s approximate posterior. The learned variances within the \texttt{PVL Adapters} capture aleatoric uncertainty, while Monte Carlo sampling accounts for epistemic uncertainty arising from model variability. The predictive entropy across sampled outputs quantifies total uncertainty and overall confidence. Empirically, we found 30 forward passes sufficient to obtain stable uncertainty estimates.

\medskip
\parhead{Residual gating:} Directly relying on attended features from \texttt{PVL Adapters} can introduce instability early in the training schedule, when attention responses are still noisy. The residual gate mitigates this by controlling how much new information is incorporated, gradually increasing reliance on attended features as cross-modal alignment and model confidence improve, leading to smoother optimization and more reliable fusion:

\begin{equation}
O_{\mathrm{proj}} = O \cdot W_{\mathrm{out}},
\end{equation}
where $W_{\mathrm{out}} \in \mathbb{R}^{D_a \times D_s}$ is the output projection matrix.

\medskip

\noindent Then we apply a learnable gate $g \in [0,1]$:
\begin{equation}
Y = g \odot O_{\mathrm{proj}} + (1 - g) \odot X,
\end{equation}
\noindent where scalar $g$ is initialized to a balanced weighting through $\texttt{sigmoid}(0)$, providing equal emphasis on the original query $X$ and the attended output at the start of training.

\medskip
\parhead{Bidirectional Interaction:} A two-way Transformer layer performs mutual updates via $\texttt{Attn}_{\texttt{PVL}}$, enabling visual and textual features to mutually refine each other for stronger cross-modal alignment and contextual consistency:
\begin{align}
\mathbf{v}'^{(n)} &= \texttt{Attn}_{\texttt{PVL}}^{v\rightarrow t}(\mathbf{v}^{(n)}, \mathbf{t}^{(n)}) \\
\mathbf{t}'^{(n)} &= \texttt{Attn}_{\texttt{PVL}}^{t \rightarrow v}(\mathbf{t}^{(n)}, \mathbf{v}'^{(n)})
\end{align}

\medskip
\parhead{Upward Projection:} Finally, the fused features are projected back to their original dimensions with $W^{\uparrow^{(n)}}_v$ $\in$ $\mathbb{R}^{D_s \times D_v}$ and $W^{\uparrow^{(n)}}_t$ $\in$ $\mathbb{R}^{D_s \times D_t}$ with residual connections applied:
\begin{align}
\hat{\mathbf{V}}^{(n)} &= \mathbf{V}^{(n)} + \mathbf{v}'^{(n)} W^{\uparrow^{(n)}}_v \\
\hat{\mathbf{T}}^{(n)} &= \mathbf{T}^{(n)} + \mathbf{t}'^{(n)} W^{\uparrow^{(n)}}_t
\end{align}

\noindent The \texttt{PVL Adapters} is applied at multiple encoding CLIP layers to refine joint representations.

\medskip

\subsection{Segmentation via Pixel-Text Similarity} After the final fusion layer, the text \texttt{[EOS]} token embedding and the visual patch are L2-normalized. The visual patch tokens are then upscaled via a learned block $\psi$, while a lightweight MLP mask head $\phi$ maps $\bm{Z_t}^{\texttt{[EOS]}}$ to a compatible embedding:
\begin{equation}
\tilde{\mathbf{V}} = \psi(\bm{Z_v}^{\texttt{[Patches]}}), \quad \tilde{\mathbf{t}} = \phi(\bm{Z_t}^{\texttt{[EOS]}})
\end{equation}

\noindent The segmentation logits $\mathbf{M} \in \mathbb{R}^{B \times H \times W}$ are computed by a simple dot product followed by bilinear interpolation:  
\begin{equation}
\mathbf{{M}} = \texttt{Upsample}_{H \times W}(\mathbf{\tilde{\mathbf{V}} \cdot \tilde{\mathbf{t}}^\top})
\end{equation}

\subsection{Soft Patch-level Contrastive Loss}

Building on CLIP’s ability to capture rich image–text relationships, we extend it to handle diverse descriptions with spatial context for medical images. Since a single caption may mention multiple anatomical regions, global alignment alone can be insufficient for fine-grained correspondence. To address this, we average image patch embeddings into stable regional representations that preserve local visual semantics while reducing token-level noise. This enables more accurate text–region associations and consistent segmentation performance across heterogeneous anatomy. Specifically, we align L2-normalized \emph{Average-pooled} visual patch embeddings $p_v = \bm{\bar{Z}_v}^{\texttt{[Patches]}}$ with text embeddings $p_t = \bm{Z_t}^{\texttt{[EOS]}}$ via a \textit{soft contrastive loss}, computing text-to-image similarity logit $\mathrm{P}^{v \rightarrow t}$ and image-to-text counterpart $\mathrm{P}^{t \rightarrow v}$ across all batch pairs.

\medskip

\noindent Since text prompts within a batch can be similar, hard targets are replaced with soft targets derived from text similarities:  
\begin{equation}
\mathrm{G} = \texttt{softmax}\left(\frac{p_t \cdot p_t^\top}{\tau}\right),
\end{equation}
\noindent where $\tau$ is a temperature parameter set to 0.2. The soft cross-entropy loss is defined in the following, and the final contrastive loss averages both directions:
\begin{equation}
\mathcal{L}_{\mathrm{soft}}(\mathrm{P}, \mathrm{G}) = -\frac{1}{B} \sum_i \sum_j \mathrm{G}_{ij} \log( \texttt{softmax}(\mathrm{P}_i)_{j}).
\end{equation}
\begin{equation}
\mathcal{L}_{\mathrm{SoftCon}} = \frac{1}{2}\left(\mathcal{L}_{\mathrm{soft}}(\mathrm{P}^{t \rightarrow v}, \mathrm{G}) + \mathcal{L}_{\mathrm{soft}}(\mathrm{P}^{v \rightarrow t}, \mathrm{G}^\top)\right).
\end{equation}

\noindent The overall training loss combines segmentation (Dice+BCE, equal weights) and contrastive objectives:
\begin{equation}
\mathcal{L} = \lambda_{\mathrm{Seg}} \, \mathcal{L}_{\mathrm{Seg}} + \lambda_{\mathrm{SoftCon}} \, \mathcal{L}_{\mathrm{SoftCon}}.
\end{equation}

\noindent where \(\lambda_{\mathrm{Seg}}, \lambda_{\mathrm{SoftCon}}\) control the relative importance of each loss term, set to 0.5 and 0.1, respectively.

%% file: sec/3_results.tex
\section{Experiments and Results}
\label{sec:experiments}

\subsection{Experimental Setup}
\label{subsec:experimental-setup}
\parhead{Data Efficiency:}  
We evaluate \texttt{MedCLIPSeg} under varying levels of training data to assess data-efficiency and scalability. Models are trained with 10\%, 25\%, and 50\% of the data to measure performance under limited supervision. The fully supervised setting uses all training examples with their pixel-level and textual annotations, serving as an upper-bound reference.

\smallskip
\parhead{Domain Generalization:}  
To assess out-of-distribution (OOD) generalization, models are trained on one in-distribution (ID) source dataset, fully supervised, and directly tested on unseen target datasets without fine-tuning. This evaluates robustness to domain shifts in image acquisition, clinical sites, and patient populations.

\input{tables/main_results_efficiency}
\input{tables/domain_generalization}

\smallskip
\parhead{Datasets:}  
We experiment on diverse medical imaging datasets spanning six organs and five modalities, covering clinically critical segmentation tasks including \emph{tumor}, \emph{polyp}, and \emph{skin lesion} segmentation, each representing challenging, high-impact applications with ambiguous boundaries and large domain variability. \textbf{For supervised settings and data-efficiency tests}, we use BUSI~\cite{busi}, BTMRI~\cite{Cheng2017}, ISIC~\cite{isic1,isic2}, Kvasir-SEG~\cite{jha2019kvasir}, QaTa-COV19~\cite{qatacov19}, and EUS~\cite{jaramillo2020endoscopic}.  
\textbf{For domain generalization tests}, evaluations are conducted on BUSUC~\cite{busuc}, BUSBRA~\cite{busbra}, BUID~\cite{buid}, UDIAT~\cite{udiat}, CVC-ColonDB~\cite{colondb}, CVC-ClinicDB~\cite{clinicdb}, CVC-300~\cite{cvc300}, BKAI~\cite{bkai}, BRISC~\cite{brisc}, and UWaterlooSkinCancer~\cite{waterloo1,waterloo2}, each introducing substantial domain shifts. For datasets with missing text, we generated prompts using GPT-5 \cite{gpt5} and applied image processing techniques to identify different visual attributes for each image. Dataset and prompt details are provided in Appendices \ref{appx:datasets_overview} and \ref{appx:text_prompt_generation}.

\smallskip
\parhead{Implementation Details:}  
We use UniMedCLIP ViT-B/16~\cite{khattak2024unimed} with PubMedBERT~\cite{biomedclip} as the backbone. All models are trained for 100~epochs (10 for EUS) with a learning rate of $3\times10^{-4}$, batch size 24, and Adam optimizer~\cite{kingma2017adam} under cosine annealing. The segmentation loss combines Dice and binary cross-entropy (BCE) losses (equal weights). Experiments are conducted on a single NVIDIA A100 GPU (40GB RAM). All settings were identical across all experiments, and all CLIP-based baselines use the same UniMedCLIP backbone for fair comparison. We use Dice Similarity Coefficient (DSC) and normalized surface distance (NSD) to compare segmentation accuracy.

% -----------------------------------------------------------------------

\subsection{Data Efficiency Evaluation}
As shown in Table~\ref{tab:main_results_efficiency}, \texttt{MedCLIPSeg} consistently outperforms both unimodal and multimodal baselines, notably the strong CLIP-based CAT-Seg, with \textbf{2--3\%} gains at 10\% data and \textbf{3--4\%} at 50\%. Compared to EoMT-CLIP~\cite{kerssies2025your}, a variant without the \texttt{PVL Adapters} and $\mathcal{L}_{\text{SoftCon}}$, \texttt{MedCLIPSeg} achieves further \textbf{+7.0\%} and \textbf{+8.8\%} DSC improvements at 10\% and 25\% efficiency, respectively, highlighting the effectiveness of these components for data-efficient learning.

% -----------------------------------------------------------------------

\subsection{Domain Generalization}
Table~\ref{tab:domain_generalization} reports cross-dataset performance. \texttt{MedCLIPSeg} achieves 85.7\% DSC on BUSI, 84.4\% on BUSUC, 90.2\% on Kvasir-SEG, 88.0\% on BTMRI, and 92.5\% on ISIC, consistently outperforming SAN~\cite{xu2023side} and CAT-Seg~\cite{cho2024cat}. Despite substantial distribution shifts, such as lighting/signal gain, zoom, and viewpoint/field-of-view variations in polyp and ultrasound datasets, our model preserves contour fidelity and segmentation quality, demonstrating robust generalization across domains.
\input{tables/component_ablation}

\input{tables/text_prompt_ablation}
% -----------------------------------------------------------------------

\begin{figure*}[!htbp]
    \centering
    \includegraphics[width=0.96\linewidth, height=0.33\linewidth]{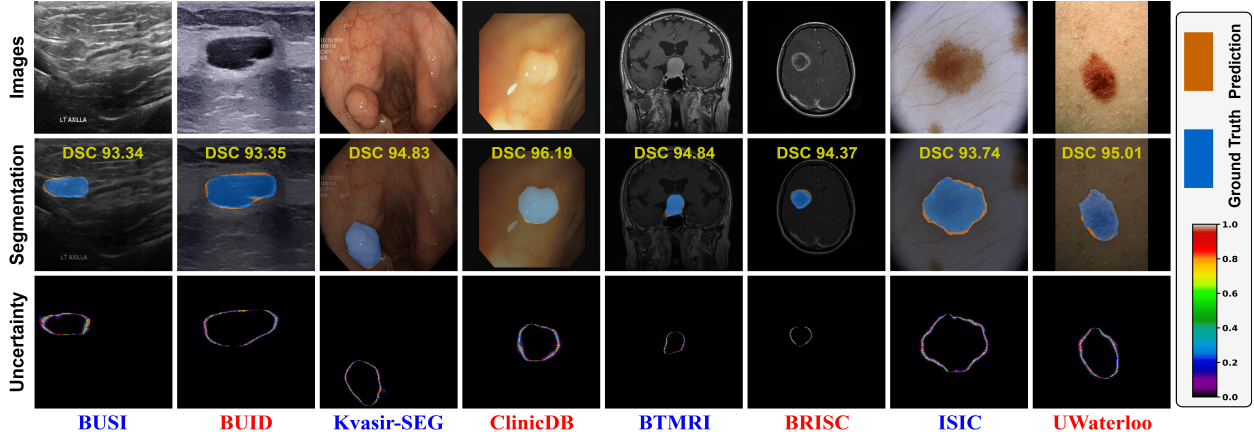}
    \caption{\textbf{Segmentation and uncertainty visualizations.} Uncertainty peaks along lesion boundaries and remains consistent across diverse datasets, indicating reliable calibration and generalization. ID data are in {\color[HTML]{0000FF}\textbf{blue}} while OOD data are in {\color[HTML]{FF0000}\textbf{red}}.}
    \label{fig:qualitative-examples}
\end{figure*}

\subsection{Effectiveness of Key Design Components}
Table~\ref{tab:component_ablation} quantifies the impacts of \texttt{MedCLIPSeg}'s components. HM denotes the harmonic mean between the ID and OOD DSC (\%) scores. Removing the \texttt{PVL Adapters} causes the largest DSC drop (–7.9\% ID, –23.8\% OOD), emphasizing their role in robust multi-modal alignment. Replacing probabilistic attention with deterministic attention reduces OOD DSC by 15.9\%, confirming the value of uncertainty-aware formulation. Bidirectional interaction further enhances performance, while excluding $\mathcal{L}_{\mathrm{SoftCon}}$ decreases HM DSC by 1.92\%, highlighting its positive role in maintaining nuanced cross-modal alignment.

% -----------------------------------------------------------------------
\begin{figure}[!htbp]
    \centering
    \includegraphics[width=0.98\linewidth, height=0.52\linewidth]{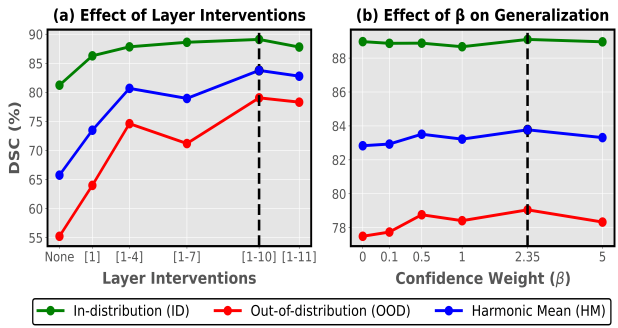}
    \caption{Layer-wise interventions (\textbf{\textit{left}}) and confidence weighting ($\beta$) (\textbf{\textit{right}}) ablations averaged on ID and OOD data.}
    \label{fig:layers_beta_plot}
\end{figure}

\input{tables/clip_model_ablation}

\subsection{Ablation Studies}
\label{subsec:ablation-studies}
\parhead{Layer Interventions:}  
Figure~\hyperref[fig:layers_beta_plot]{\ref*{fig:layers_beta_plot}(a)} shows that deeper layer interventions with \texttt{PVL Adapters} steadily improve both ID and OOD segmentation, peaking at Layer~10 (HM DSC 83.76\%), with a slight drop at the final layer.

\vspace{1.6pt}
\parhead{Confidence Weight ($\beta$) Choice:}  
Figure~\hyperref[fig:layers_beta_plot]{\ref*{fig:layers_beta_plot}(b)} shows that $\beta{=}2.35$ yields the best HM DSC from a range of tested values in [0,5], balancing in-domain stability and OOD generalization by effectively calibrating probabilistic attention.

\vspace{1.6pt}
\parhead{Text Prompt Style:}  
% As \texttt{MedCLIPSeg} uses descriptive text to drive segmentation,
Table~\ref{tab:prompt-config} shows that concise yet informative prompts (i.e., \texttt{Original}) perform best. Removing spatial cues (i.e., location) or adding verbosity lowers performance by 1.65\% and 5.28\%, respectively. \texttt{Contradictory}, which contain internally inconsistent descriptions that can confuse the model, along with \texttt{Underspecified} prompts, deteriorate DSCs to 65.79\% and 56.82\%, respectively (see Appendix \ref{appx:prompt_configs} for more details). 

% Prompts that clearly anchor anatomical regions help the model align visual and textual cues, whereas overly long or ambiguous phrasing dilutes the attention focus.  A qualitative overview of these prompt variations is provided in the \textit{Supplementary Materials}.  

\vspace{1.6pt}
\parhead{CLIP Backbone:}  
Table~\ref{tab:clip_model_ablation} shows that the backbone choice strongly impacts generalization. Models benefit from stronger and larger-scale CLIP pretraining, with UniMedCLIP \cite{khattak2024unimed} providing the most robust transferability.

\subsection{Uncertainty Visualization and Reliability}
\label{sec:qualitative}
Figure~\ref{fig:qualitative-examples} illustrates segmentation and uncertainty maps for both ID and OOD data, where uncertainty consistently concentrates along anatomical boundaries and regions prone to expert disagreement. All subsequent quantitative analyses are computed over \textbf{foreground regions} in ID and OOD data. Predicted uncertainty closely follows segmentation errors, achieving strong Spearman correlations of \textbf{(87.57\%, 80.41\%)} for (ID, OOD), respectively. Furthermore, probabilistic modeling improves calibration, reducing the Brier scores from \textbf{(23.9\%, 25.3\%)} in the deterministic baseline to \textbf{(11.1\%, 11.8\%)}, as shown in Fig.~\ref{fig:figure1}, alleviating overconfidence and enhancing reliability in clinical decision-making.

% -----------------------------------------------------------------------

%% file: tables/main_results_efficiency.tex
\begin{table}[!htbp]
\centering
\caption{\textbf{Data-efficiency evaluation:} 
This table reports the average DSC and NSD (\%) when varying the fraction of training data across different segmentation methods. Best results are in \textbf{bold}, and second-best are \underline{underlined}.}
\tablestyle{-12pt}{1.1}
\addtolength{\tabcolsep}{+14pt}
\resizebox{0.98\linewidth}{!}{%
\begin{tabular}{ccccccccc}
\toprule
\multicolumn{1}{c}{\multirow{2}{*}{\textbf{Method}}} &
\multicolumn{2}{c}{\textbf{10\% Data}} &
\multicolumn{2}{c}{\textbf{25\% Data}} &
\multicolumn{2}{c}{\textbf{50\% Data}} &
\multicolumn{2}{c}{\textbf{100\% Data}} \\
\cmidrule(lr){2-3}
\cmidrule(lr){4-5}
\cmidrule(lr){6-7}
\cmidrule(lr){8-9}
& \textbf{DSC} $\uparrow$ & \textbf{NSD} $\uparrow$
& \textbf{DSC} $\uparrow$ & \textbf{NSD} $\uparrow$
& \textbf{DSC} $\uparrow$ & \textbf{NSD} $\uparrow$
& \textbf{DSC} $\uparrow$ & \textbf{NSD} $\uparrow$ \\
\midrule
\rowcolor{gray!15} \multicolumn{9}{c}{\texttt{\textbf{Unimodal Approaches}}} \\
UNet \cite{ronneberger2015unet} & 60.95 & 64.43 & 62.74 & 66.16 & 71.61 & 75.14 & 78.49 & 82.07 \\
UNet++ \cite{zhou2018unet++} & 63.72 & 67.08 & 65.86 & 69.21 & 73.15 & 76.31 & 78.44 & 81.79 \\
DeepLabv3 \cite{chen2017rethinking} & 61.32 & 64.84 & 65.39 & 69.10 & 68.58 & 72.57 & 73.28 & 77.42 \\
AttnUNet \cite{oktay2018attention} & 62.78 & 66.25 & 64.97 & 68.53 & 71.34 & 74.96 & 76.30 & 79.77 \\
nnUNet \cite{isensee2021nnu} & 73.45 & 77.37 & 76.73 & 80.66 & 78.86 & 82.68 & 81.40 & 85.08 \\
Swin-UNet \cite{cao2022swin} & 53.04 & 57.91 & 54.69 & 59.24 & 55.89 & 61.25 & 65.03 & 69.32 \\
TransUNet \cite{chen2021transunet} & 52.69 & 56.38 & 55.25 & 58.95 & 55.22 & 59.30 & 67.22 & 71.15 \\ \midrule
\rowcolor{gray!15} \multicolumn{9}{c}{\texttt{\textbf{Generic Text-driven Approaches}}} \\
% LAVT \cite{yang2022lavt} & 68.66 & 71.08 & 72.45 & 74.99 & 79.07 & 81.71 & 82.17 & 84.84 \\
LViT \cite{li2023lvit} & 66.51 & 68.80 & 75.66 & 78.12 & 78.88 & 81.34 & 83.35 & 85.89 \\
Ariadne's Thread \cite{zhong2023ariadne} & 61.34 & 62.75 & 63.09 & 64.51 & 65.65 & 66.92 & 70.07 & 71.49 \\ \midrule
\rowcolor{gray!15} \multicolumn{9}{c}{\texttt{\textbf{CLIP-Based Approaches}}} \\ 
EoMT-CLIP \cite{kerssies2025your} & 74.07 & 77.41 & 76.29 & 79.84 & 79.19 & 82.78 & 82.93 & 86.35 \\
CLIPSeg \cite{luddecke2022image} & 74.66 & 77.75 & 78.31 & 81.34 & 79.63 & 82.58 & 84.87 & 87.74 \\
DenseCLIP \cite{rao2022denseclip} & 67.84 & 70.33 & 70.23 & 72.70 & 72.09 & 74.45 & 74.19 & 76.89 \\
ZegCLIP \cite{zhou2023zegclip} & 61.25 & 63.72 & 72.46 & 75.01 & 76.21 & 78.80 & 78.98 & 81.69 \\
SAN \cite{xu2023side} & 74.13 & 76.97 & 76.13 & 78.91 & 78.80 & 81.52 & 81.62 & 84.35 \\
MaPLe \cite{khattak2023maple} & 66.27 & 68.75 & 71.53 & 73.95 & 74.60 & 77.12 & 74.60 & 77.10 \\
MaPLe \cite{khattak2023maple} + Decoder & 74.81 & 77.90 & 79.64 & 82.60 & 82.81 & \underline{85.80} & 84.94 & 87.91 \\
VLSM-Adapter \cite{dhakal2024vlsm} & 74.47 & 77.50 & 77.63 & 80.53 & 80.83 & 83.77 & 83.85 & 86.72 \\
CausalCLIPSeg \cite{chen2024causalclipseg} & 71.19 & 73.74 & 75.42 & 78.00 & 78.60 & 81.22 & 81.34 & 84.20 \\
CAT-Seg \cite{cho2024cat} & \underline{78.76} & \underline{81.50} & \underline{81.12} & \underline{83.92} & \underline{83.32} & 85.61 & \underline{85.90} & \underline{88.31} \\
\midrule
\rowcolor{lightpurple}
\texttt{MedCLIPSeg} (Ours) & \textbf{81.10} & \textbf{83.94} & \textbf{85.08} & \textbf{87.85} & \textbf{87.18} & \textbf{89.95} & \textbf{88.66} & \textbf{91.35} \\
\bottomrule
\end{tabular}
}
\label{tab:main_results_efficiency}
\end{table}

%% file: tables/domain_generalization.tex
\begin{table*}[ht]
\centering
\caption{\textbf{Domain generalization:} 
Models are trained on a source dataset and evaluated on OOD target datasets without adaptation. DSC (\%) values are reported where the best results are in \textbf{bold}, and second-best are \underline{underlined}.}
\setlength{\tabcolsep}{4pt}
\resizebox{0.98\linewidth}{!}{%
\begin{tabular}{l ccccc ccccc cc cc}
\toprule
{\multirow{4}{*}{\textbf{Method}}} & \multicolumn{5}{c}{\textbf{Breast Ultrasound}} 
& \multicolumn{5}{c}{\textbf{Polyp Endoscopy}}
& \multicolumn{2}{c}{\textbf{Brain MRI}} 
& \multicolumn{2}{c}{\textbf{Skin Dermatoscopy}} \\
\cmidrule(lr){2-6} \cmidrule(lr){7-11} \cmidrule(lr){12-13} \cmidrule(lr){14-15}
& \textbf{Source} & \multicolumn{4}{c}{\textbf{Target}} 
& \textbf{Source} & \multicolumn{4}{c}{\textbf{Target}} 
& \textbf{Source} & \textbf{Target}
& \textbf{Source} & \textbf{Target} \\
\cmidrule(lr){2-2} \cmidrule(lr){3-6} \cmidrule(lr){7-7} \cmidrule(lr){8-11} \cmidrule(lr){12-12} \cmidrule(lr){13-13} \cmidrule(lr){14-14} \cmidrule(lr){15-15}
& BUSI & BUSBRA & BUSUC & BUID & UDIAT 
& Kvasir-SEG & ColonDB & ClinicDB & CVC300 & BKAI
& BTMRI & BRISC
& ISIC & UWaterloo \\
\midrule
% LAVT \cite{yang2022lavt}           & 79.26 & 48.32 & 62.52 & 5.68 & 45.26 & 74.55 & 42.25 & 57.57 & 41.72 & 53.39 & 83.38 & 67.64 & 90.44 & 23.37 \\
LViT \cite{li2023lvit}              & 75.32 & 59.41 & 67.95 & 53.51 & 65.60 & 85.29 & 60.01 & 75.27 & 70.22 & 67.17 & 81.41 & 71.86 & 91.21 & 58.87 \\
% Ariadne's Thread \cite{zhong2023ariadne} & 57.26 & 20.67 & 49.36 & 22.29 & 22.19 & 77.42 & 23.08 & 22.99 & 20.18 & 22.19 & 69.96 & 25.89 & 68.37 & 43.43 \\
CLIPSeg \cite{luddecke2022image}        & 80.95 & 63.66 & 75.03 & 68.43 & 56.67 & 81.98 & 59.93 & 71.49 & 72.74 & 66.46 & \underline{86.33} & \underline{77.61} & 90.55 & 80.19 \\
DenseCLIP \cite{rao2022denseclip}       & 71.85 & 53.34 & 70.97 & 63.53 & 54.93 & 79.32 & 56.38 & 68.08 & 64.71 & 61.63 & 70.30 & 34.12 & 89.29 & 53.39 \\
ZegCLIP \cite{zhou2023zegclip}          & 72.08 & 61.08 & 73.57 & 71.75 & 52.41 & 78.46 & 53.46 & 69.75 & 60.73 & 65.60 & 76.65 & 66.31 & 81.45 & 38.60 \\
SAN \cite{xu2023side}                   & 77.99 & 64.37 & 74.15 & 58.13 & 61.98 & 83.16 & 61.82 & 74.46 & \underline{80.36} & 69.31 & 85.27 & 71.60 & \underline{91.39} & \underline{82.51} \\
MaPLe \cite{khattak2023maple}           & 66.37 & 50.08 & 71.52 & 70.77 & 57.81 & 76.12 & 48.09 & 59.64 & 63.80 & 56.94 & 75.40 & 45.19 & 88.31 & 69.12 \\
MaPLe \cite{khattak2023maple} + Decoder & 80.49 & 55.89 & 64.96 & 60.66 & 59.44 & 83.46 & 61.53 & 71.20 & 74.62 & 66.93 & 85.08 & 71.46 & 90.10 & 81.83 \\
VLSM-Adapter \cite{dhakal2024vlsm}      & 80.90 & 68.48 & \underline{82.37} & \underline{75.26} & 69.16 & 85.89 & 63.51 & \underline{76.09} & 75.24 & 71.59 & 85.03 & 68.92 & 91.30 & 82.17 \\
CausalCLIPSeg \cite{chen2024causalclipseg} & 76.11 & 55.87 & 69.12 & 64.49 & 48.90 & 78.77 & 41.65 & 57.54 & 45.77 & 52.56 & 81.71 & 53.96 & 89.47 & 48.73 \\
CAT-Seg \cite{cho2024cat} & \underline{81.83} & \underline{70.94} & 81.48 & 73.37 & \underline{70.30} & \underline{86.43} & \underline{68.49} &  70.35 & 78.12 & \underline{74.35} & 84.86 & 76.28 & 91.27 & 82.02 \\
\midrule
\rowcolor{lightpurple} \texttt{MedCLIPSeg} (Ours)
& \textbf{85.72} & \textbf{75.06} & \textbf{84.37} & \textbf{78.99} & \textbf{74.64} 
& \textbf{90.15} & \textbf{71.90} & \textbf{80.80} & \textbf{80.82} & \textbf{79.15} 
& \textbf{88.03} & \textbf{80.92} & \textbf{92.54} & \textbf{83.53} \\
\bottomrule
\end{tabular}}
\label{tab:domain_generalization}
\end{table*}

%% file: tables/component_ablation.tex
\begin{table}[!htbp]
\caption{\textbf{Effect of different key components.}}
\tablestyle{-22pt}{1.2}
\addtolength{\tabcolsep}{+24pt}
\resizebox{0.98\linewidth}{!}{
\centering
\begin{tabular}{lccc}
\toprule
\multicolumn{1}{c}{\multirow{2}{*}{\textbf{Method}}} & 
\multicolumn{3}{c}{\textbf{Domain Generalization}} \\ 
\cmidrule(lr){2-4}
 & \textbf{ID DSC (\%)} & \textbf{OOD DSC (\%)} & \textbf{HM DSC (\%)} \\
\midrule
\rowcolor{lightpurple} \texttt{MedCLIPSeg} (Ours) & \textbf{89.11} & \textbf{79.02} & \textbf{83.76} \\
\midrule
\rowcolor{gray!15} \multicolumn{4}{c}{\texttt{\textbf{Probabilistic Vision-Language Adapters}}} \\
w/o \texttt{PVL Adapters} & 81.23$_{(-7.88)}$\textcolor{red}{$\downarrow$} & 55.23$_{(-23.79)}$\textcolor{red}{$\downarrow$} & 65.75$_{(-18.01)}$\textcolor{red}{$\downarrow$} \\
w/o Gating & 87.55$_{(-1.56)}$\textcolor{red}{$\downarrow$} & 76.79$_{(-2.23)}$\textcolor{red}{$\downarrow$} & 81.82$_{(-1.94)}$\textcolor{red}{$\downarrow$} \\
w/o  $\texttt{Attn}_{\texttt{PVL}}$  & 86.21$_{(-2.90)}$\textcolor{red}{$\downarrow$} & 74.13$_{(-4.89)}$\textcolor{red}{$\downarrow$} & 79.71$_{(-4.05)}$\textcolor{red}{$\downarrow$} \\
Deterministic \texttt{MedCLIPSeg} & 87.68$_{(-1.43)}$\textcolor{red}{$\downarrow$} & 63.12$_{(-15.90)}$\textcolor{red}{$\downarrow$} & 73.40$_{(-10.36)}$\textcolor{red}{$\downarrow$} \\
\midrule
\rowcolor{gray!15} \multicolumn{4}{c}{\texttt{\textbf{Bidirectional Multimodal Interaction}}} \\
w/o Visual Adaptation & 81.50$_{(-7.61)}$\textcolor{red}{$\downarrow$} & 64.40$_{(-14.62)}$\textcolor{red}{$\downarrow$} & 71.95$_{(-11.81)}$\textcolor{red}{$\downarrow$} \\
w/o Textual Adaptation & 88.83$_{(-0.28)}$\textcolor{red}{$\downarrow$} & 76.40$_{(-2.62)}$\textcolor{red}{$\downarrow$} & 82.15$_{(-1.61)}$\textcolor{red}{$\downarrow$} \\
w/o Bidirectional Interaction & 88.71$_{(-0.40)}$\textcolor{red}{$\downarrow$} & 77.71$_{(-1.31)}$\textcolor{red}{$\downarrow$} & 82.85$_{(-0.91)}$\textcolor{red}{$\downarrow$} \\
Unimodal \texttt{MedCLIPSeg} & 86.53$_{(-2.58)}$\textcolor{red}{$\downarrow$} & 73.49$_{(-5.53)}$\textcolor{red}{$\downarrow$} & 79.48$_{(-4.28)}$\textcolor{red}{$\downarrow$} \\
\midrule
\rowcolor{gray!15} \multicolumn{4}{c}{\texttt{\textbf{Contrastive Loss}}} \\
w/o $\mathcal{L}_{\mathrm{SoftCon}}$ & 87.24$_{(-1.87)}$\textcolor{red}{$\downarrow$} & 77.08$_{(-1.94)}$\textcolor{red}{$\downarrow$} & 81.84$_{(-1.92)}$\textcolor{red}{$\downarrow$} \\
Hard Targets & 88.34$_{(-0.77)}$\textcolor{red}{$\downarrow$} & 77.64$_{(-1.38)}$\textcolor{red}{$\downarrow$} & 82.65$_{(-1.11)}$\textcolor{red}{$\downarrow$} \\
Attention-pooled $\mathcal{L}_{\mathrm{SoftCon}}$ & 88.73$_{(-0.38)}$\textcolor{red}{$\downarrow$} & 75.60$_{(-3.42)}$\textcolor{red}{$\downarrow$} & 81.64$_{(-2.12)}$\textcolor{red}{$\downarrow$} \\
\bottomrule
\end{tabular}
}
\label{tab:component_ablation}
\end{table}

%% file: tables/text_prompt_ablation.tex
\begin{table}[!htbp]
\centering
\caption{\textbf{Effect of text prompt design.}}
\tablestyle{-12pt}{1.1}
\addtolength{\tabcolsep}{+14pt}
\resizebox{0.98\linewidth}{!}{%
\begin{NiceTabular}{c|ccc}
\toprule
 \textbf{Text Prompt Design}                   & \textbf{ID DSC (\%)}           & \textbf{OOD DSC (\%)}          & \textbf{HM DSC (\%)}    \\ \midrule
 \texttt{Contradictory} & 68.60         & 63.21         &   65.79        \\
 \texttt{Missing Location} & 86.98         & 77.75          &   82.11        \\
 \texttt{Overdescriptive} & 82.93         & 74.49          &   78.48        \\
 \texttt{Underdescriptive} & 66.91         & 49.38          &   56.82        \\
\rowcolor{lightpurple} \texttt{Original} & \textbf{89.11}          & \textbf{79.02}          &  \textbf{83.76}         \\
\bottomrule
\end{NiceTabular}%
}
\label{tab:prompt-config}
\end{table}

%% file: tables/clip_model_ablation.tex
\begin{table}[!htbp]
\centering
\caption{\textbf{Effect of pre-trained vision–language models.}}
\tablestyle{-12pt}{1.1}
\addtolength{\tabcolsep}{+14pt}
\resizebox{0.98\linewidth}{!}{%
\begin{NiceTabular}{c|ccc}
\toprule
 \textbf{Pre-trained Model}                   & \textbf{ID DSC (\%)}           & \textbf{OOD DSC (\%)} &  \textbf{HM DSC (\%)} \\ \midrule
 CLIP \cite{radford2021learning} & 88.48 & 74.81 & 81.07 \\
 PubMedCLIP \cite{eslami-etal-2023-pubmedclip} & 86.67 & 73.05 & 79.28 \\
 BiomedCLIP \cite{biomedclip} & 88.70 & 77.08 & 82.48 \\
\rowcolor{lightpurple} UniMedCLIP \cite{khattak2024unimed} & \textbf{89.11}          & \textbf{79.02}          &  \textbf{83.76}         \\
\bottomrule
\end{NiceTabular}%
}
\label{tab:clip_model_ablation}
\end{table}

%% file: sec/4_conclusion.tex
\section{Conclusion}
\label{sec:conclusion}
We presented \texttt{MedCLIPSeg}, a probabilistic framework for text-driven medical image segmentation by adopting the novel \texttt{PVL Adapter} that enables confidence-weighted attention and explicit uncertainty estimation. Leveraging CLIP’s pretrained features through bidirectional fusion and a soft patch-level contrastive loss, \texttt{MedCLIPSeg} achieves SOTA segmentation performance with high data efficiency, strong out-of-distribution generalization, and well-calibrated uncertainty across six organs and five modalities, advancing reliable VLM for medical AI.

\section*{Acknowledgements} 
We acknowledge the support of the Natural Sciences and Engineering Research Council of Canada (NSERC) and the Fonds de recherche du Qu\'ebec – Nature et technologies (B2X-363874).

%% file: sec/X_suppl.tex
\clearpage
\renewcommand{\thesection}{\Alph{section}}
\setcounter{section}{0}
\renewcommand{\thetable}{S\arabic{table}}
\renewcommand{\thefigure}{S\arabic{figure}}
\setcounter{table}{0}
\setcounter{figure}{0}
\renewcommand{\theHsection}{supp-\arabic{section}}
\maketitlesupplementary

\section{Datasets Overview}
\label{appx:datasets_overview}
Our evaluation comprises a diverse collection of medical image segmentation datasets, covering six organs and five imaging modalities. We organize our benchmarks into three settings: \textit{data efficiency}, \textit{fully supervised}, and \textit{domain generalization}. The \textbf{data efficiency} and \textbf{fully supervised} evaluation includes BUSI~\cite{busi}, BTMRI~\cite{Cheng2017}, ISIC~\cite{isic1,isic2}, Kvasir-SEG~\cite{jha2019kvasir}, QaTa-COV19~\cite{qatacov19}, and EUS~\cite{jaramillo2020endoscopic}, which collectively span ultrasound, MRI, dermatoscopy, endoscopy, and X-ray modalities. For \textbf{domain generalization}, we employ ten diverse datasets to provide out-of-domain (OOD) samples: BUSUC~\cite{busuc}, BUSBRA~\cite{busbra}, BUID~\cite{buid}, UDIAT~\cite{udiat}, BRISC~\cite{brisc}, UWaterlooSkinCancer~\cite{waterloo1,waterloo2}, CVC-ColonDB~\cite{colondb}, CVC-ClinicDB~\cite{clinicdb}, CVC-300~\cite{cvc300}, and BKAI~\cite{bkai}, each introducing distinct appearance shifts across imaging devices, acquisition protocols, and anatomical domains. This combination enables a systematic analysis of segmentation robustness across both intra- and cross-domain distributions. Dataset statistics, modalities, and split details are summarized in Table~\ref{tab:datasets}. Importantly, our method does not use the validation sets; however, other methods, such as LViT \cite{li2023lvit}, rely on them during training to select the best checkpoints. In our framework, models are trained on the training split, and we select the last epoch checkpoint to evaluate on the test split. For \textbf{domain generalization}, the OOD datasets are \textit{never seen during training}; we evaluate directly on their test sets without any finetuning or adaptation.
\input{tables/dataset_details}

\section{Computational Cost Analysis}
Table~\ref{tab:computational-cost} summarizes the computational complexity of all compared methods, including parameter footprint, FLOPs, and inference time. All measurements are computed on the same BUSI \cite{busi} test set under identical hardware conditions. Although our \texttt{MedCLIPSeg} framework typically employs a sampling strategy during inference, the computational cost reported in Table~\ref{tab:computational-cost} corresponds to the configuration where we use \textit{only a single sampled forward pass}. This ensures a fair, per-sample comparison with other methods. In general, \texttt{MedCLIPSeg} exhibits a fair, competitive computational profile with state-of-the-art segmentation performance.
\input{tables/computational_cost}

\section{Text Prompt Generation}
\label{appx:text_prompt_generation}
We introduce a scalable strategy for \textbf{automated caption generation in unpaired datasets} without relying on vision--language models, detailed in Algorithm~\ref{alg:text_prompt_generation}. Instead of requiring image--text pairs, we query a large language model \textit{once per dataset} to produce a small set of generic caption templates with placeholders for attributes such as \textit{class, location, number, shape}, and \textit{color}. Using lightweight image and mask processing, these attributes are automatically extracted and filled into the templates, enabling ``clinician-style'' descriptive captions.
\input{tables/text_prompt_generation_psuedocode}

\section{Detailed Hyperparameters}
All models were trained using UniMedCLIP ViT-B/16 \cite{khattak2024unimed} as the vision backbone and PubMedBERT \cite{biomedclip} as the text encoder. We employed the Adam \cite{kingma2017adam} optimizer with a learning rate of $3\times10^{-4}$, a batch size of $24$, and a cosine annealing learning rate schedule. 
The segmentation objective combines Dice and binary cross-entropy losses with equal weighting ($\lambda_{Seg}\mathcal{L}_{Seg} = \lambda_{Seg}\mathcal{L}_{Dice} + \lambda_{Seg}\mathcal{L}_{BCE}$ with $\lambda_{\mathrm{Seg}} = 0.5$), 
while the CLIP-based contrastive alignment term was weighted by $\lambda_{\mathrm{SoftCon}} = 0.1$. 
The probabilistic attention weighting factor was fixed at $\beta = 2.35$ across all experiments. 
All runs were performed on a single NVIDIA A100 GPU (40~GB). Due to the relatively large size of the EUS dataset, we observed a convergence within the first 10 epochs. Consequently, EUS experiments were limited to $10$~epochs, whereas all other datasets were trained for $100$~epochs to ensure full convergence under both data-efficient and domain-generalization settings. No validation set was used, and the checkpoint at the last epoch was utilized.
\section{Prompt Designs Overview}
\label{appx:prompt_configs}
\label{sec:prompt-design-overview}
Table~\ref{tab:prompt-design-examples} provides an overview of the text prompt configurations used in our ablation experiments (see Section~\ref{subsec:ablation-studies}). Each design type represents a distinct linguistic variation that probes the model’s sensitivity to descriptive accuracy, spatial specificity, verbosity, and potential contradictions. By comparing these designs, we evaluate how differences in text formulation, from concise and spatially informative prompts to noisy or underspecified ones, influence segmentation performance and generalization across datasets.

\input{tables/prompt_design_overview}

\section{3D applicability}
\texttt{MedCLIPSeg} naturally extends to 3D segmentation without modifying the core method, when given a 3D VLM backbone. We showcase this by using M3D-CLIP \cite{bai2024m3d} to replace the 2D image encoder with a 3D one. 3D segmentations are obtained by computing the dot product between the global text token and 3D voxel features, followed by trilinear interpolation for upscaling. We validate this on the \textit{CHAOS CT Liver dataset} \cite{KAVUR2021101950} following the M3D-Seg data split and achieve a DSC of \textbf{88.72\%}, demonstrating that \texttt{MedCLIPSeg} generalizes beyond 2D. We report the \textit{average runtime per volume} over 20 test volumes in the last column of Table \ref{tab:resource_sampling}, confirming practical feasibility for 3D settings, with further 3D analysis left for future work.

\input{tables/sampling_cost}
\section{Inference-time MC sampling cost}
Table~\ref{tab:resource_sampling} shows that using 5–10 MC samples only marginally affects DSC and uncertainty estimates. 2D runtimes are reported as the \textit{average inference time per batch of 32 images}, measured over 1,000 test images, all on a single NVIDIA A100 GPU (40GB RAM). This demonstrates suitability for practical clinical settings with substantially reduced computational cost compared to 30 MC samples. In endoscopic video settings requiring $\sim$25–30 FPS, a 5 MC samples configuration is practical for real-time inference.

\section{Effect of $\lambda_{SoftCon}$}
\begin{figure}
    \centering
    \includegraphics[width=0.98\linewidth]{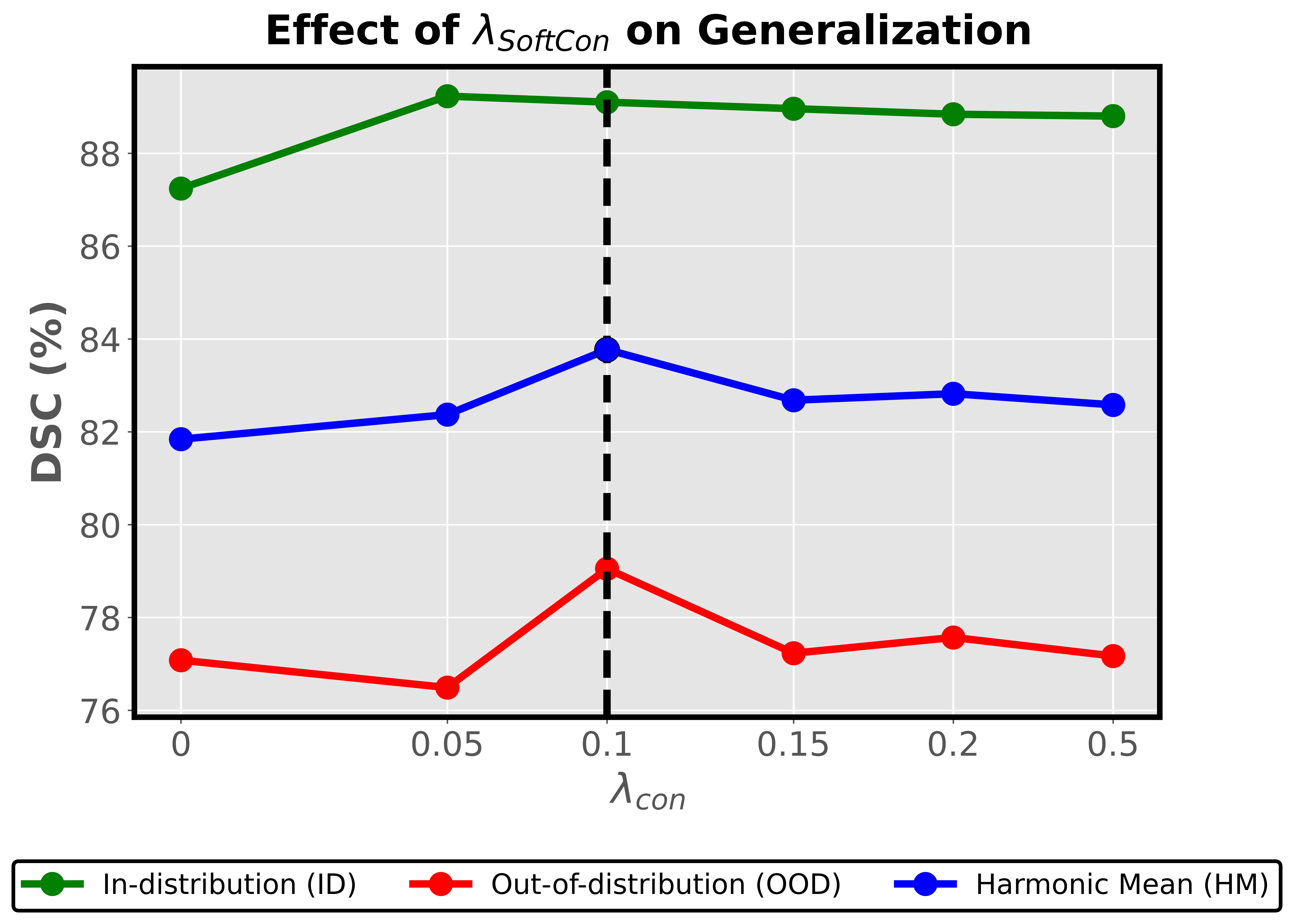}
    \caption{\textbf{Effect of $\lambda_{SoftCon}$ on Domain Generalization}}
    \label{fig:loss-con}
\end{figure}

Figure~\ref{fig:loss-con} illustrates the effect of the patch-level contrastive weight $\lambda_{SoftCon}$ on domain generalization. We find that $\lambda_{SoftCon}=0.1$ provides the optimal balance between in-distribution (ID) and out-of-distribution (OOD) performance, yielding the highest harmonic mean (HM) score. When $\lambda_{SoftCon}=0$, the contrastive loss is removed entirely, leading to a degradation in both ID and OOD performance due to the absence of semantic alignment across image–text patches. Increasing $\lambda_{SoftCon}$ beyond 0.1 results in marginal performance drops, suggesting that excessive contrastive weighting can overconstrain the feature space. These results highlight the importance of moderate patch-level contrastive regularization in maintaining both semantic consistency and domain robustness.

\section{Effect of Gating Initialization}
Table~\ref{tab:gating-init} examines the impact of the gating parameter initialization on segmentation performance across in-distribution (ID) and out-of-distribution (OOD) domains. We observe that initializing the gate with $\texttt{sigmoid(0)}$ yields the best overall results, achieving the highest ID, OOD, and harmonic mean (HM) DSC scores. A smaller initialization value (\texttt{sigmoid(-0.5)}) makes the gate overly suppressive early in training, limiting information flow from the probabilistic branch and reducing generalization. Conversely, a larger initialization (\texttt{sigmoid(0.5)}) biases the fusion toward the probabilistic output too soon, leading to mild overfitting. The balanced initialization at $\texttt{sigmoid(0)}$ thus provides a stable midpoint, enabling adaptive modulation between deterministic and probabilistic pathways throughout training.
\input{tables/gating_init}

\section{Effect of ``Two-way'' Mechanism}
Table~\ref{tab:two-way} evaluates the contribution of the two-way cross-modal attention mechanism to segmentation performance. In the \texttt{Vision First} variant, cross-modal features are first computed for the vision tokens as the query in $\texttt{Attn}_{\texttt{PVL}}$ and then refined in the subsequent text-to-image interaction, while in \texttt{Text First}, this order is reversed. Among these, initializing fusion with the visual stream (\texttt{Vision First}) achieves the best results across in-distribution (ID), out-of-distribution (OOD), and harmonic mean (HM) DSC scores. Removing the two-way mechanism (\texttt{None}) or prioritizing text-driven conditioning (\texttt{Text First}) both lead to noticeable drops in OOD generalization, indicating that early visual grounding provides a stronger foundation for subsequent text-guided refinement. This suggests that vision-first bidirectional fusion promotes more stable multimodal alignment, allowing the model to capture anatomical context before integrating semantic cues from text.
\input{tables/twoway}

\section{Effect of Contrastive Pooling Mechanism}
Table~\ref{tab:con-loss} analyzes the impact of different pooling strategies used in the contrastive loss. Among the three variants, \texttt{Average Pooling} achieves the highest in-distribution (ID), out-of-distribution (OOD), and harmonic mean (HM) DSC scores. This indicates that averaging patch-level embeddings provides a more balanced and stable global representation for contrastive learning compared to \texttt{[CLS]} or attention-based pooling. Removing uniform averaging (as in \texttt{Attention Pooling}) leads to noisier supervision due to bias toward high-attention regions, while relying solely on the \texttt{[CLS]} token underutilizes spatial information critical for dense prediction. Thus, average pooling yields the most consistent global-text alignment and best domain generalization.

\input{tables/contrastive_loss}

\section{Effect of Upscaling Blocks}
Table~\ref{tab:upscale-blocks} examines how varying the number of upscaling layers in the decoder affects segmentation performance. Using two upscaling blocks yields the best balance across in-distribution (ID), out-of-distribution (OOD), and harmonic mean (HM) DSC scores. A single block (\texttt{1}) limits spatial resolution recovery, resulting in coarse boundary predictions, while deeper configurations (\texttt{3}) introduce over-smoothing and reduce OOD robustness. The two-block design thus offers the optimal trade-off between preserving fine structural details and maintaining stable feature generalization across domains.
\input{tables/upscaling_blocks}

\section{Effect of Adapter Dimension ($D_s$)}
Table~\ref{tab:adapter-dim} evaluates the impact of the shared dimensionality in the probabilistic vision-language (PVL) adapters. The best performance is achieved at a dimension of 256, balancing both in-distribution (ID) and out-of-distribution (OOD) segmentation accuracy. Smaller adapter sizes (e.g., 64 or 128) underfit the cross-modal representations, limiting their ability to capture nuanced semantic alignments between visual and textual features. Conversely, excessively large dimensions (e.g., 512) tend to overfit the training distribution, slightly reducing OOD generalization. The 256-dimensional configuration thus provides the optimal trade-off between expressiveness and regularization.
\input{tables/adapter_dim}

\begin{figure*}
    \centering
    \includegraphics[width=0.85\linewidth, height=0.45\linewidth]{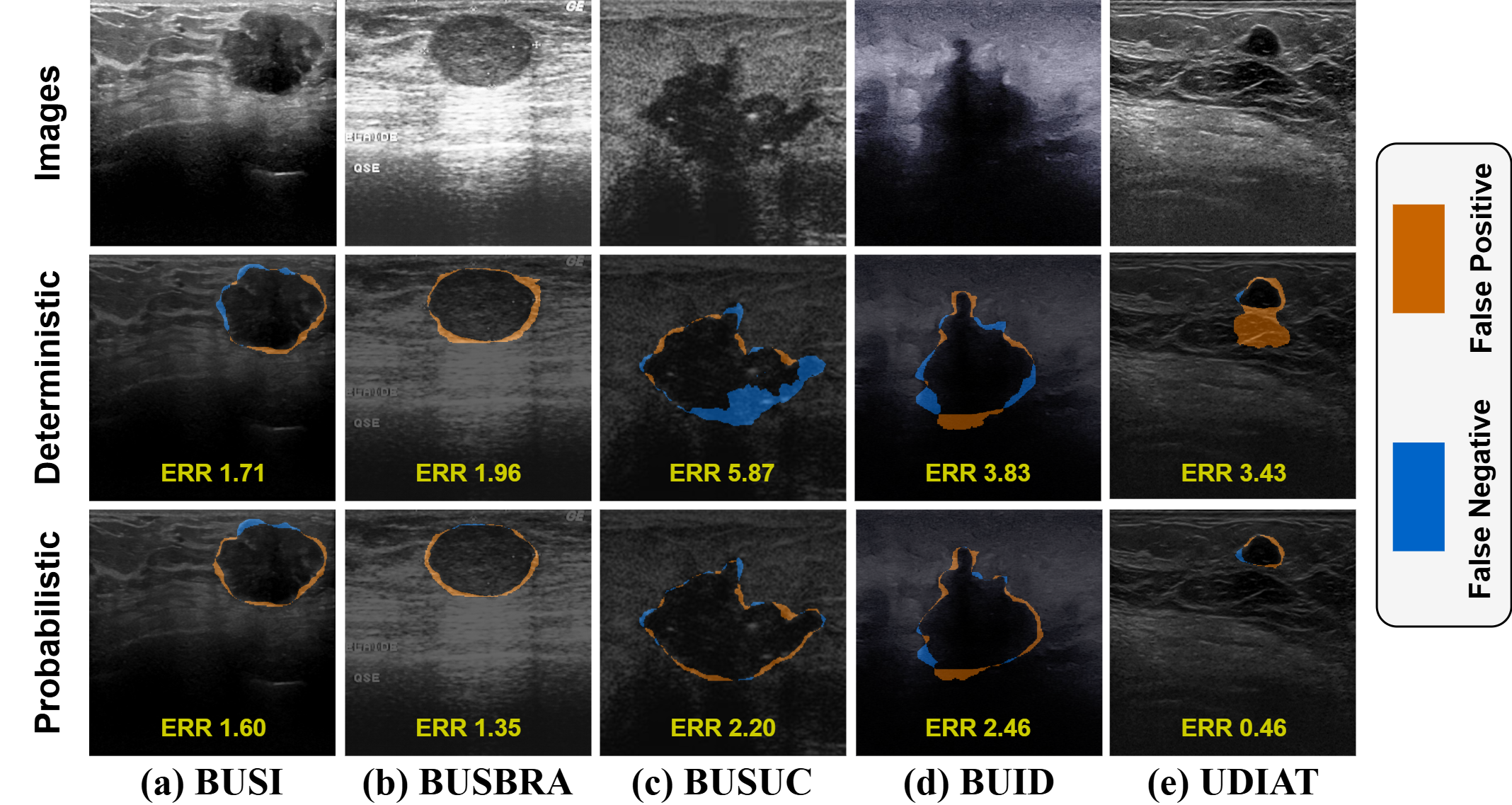}
    \caption{\textbf{FP/FN comparison between deterministic and probabilistic \texttt{MedCLIPSeg}.}}
    \label{fig:conf-examples}
\end{figure*}

\begin{figure*}
    \centering
    \includegraphics[width=0.85\linewidth, height=0.45\linewidth]{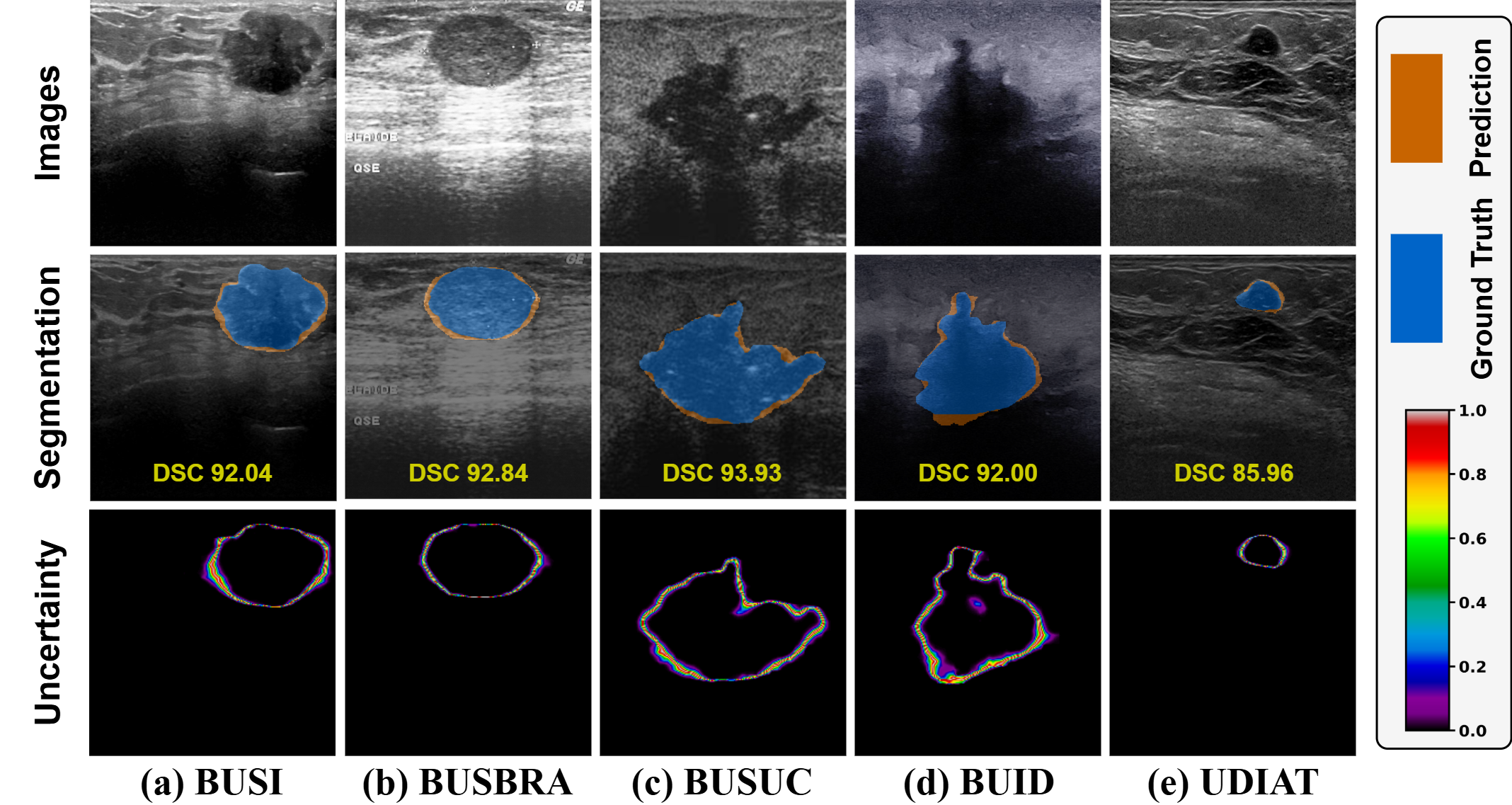}
    \caption{\textbf{Breast Tumor Ultrasound Segmentation and uncertainty visualizations.} Uncertainty peaks along lesion boundaries and remains consistent across breast ultrasound datasets, indicating reliable calibration and generalization.}
    \label{fig:segexamples-breastus}
\end{figure*}

\begin{figure*}
    \centering
    \includegraphics[width=0.85\linewidth, height=0.45\linewidth]{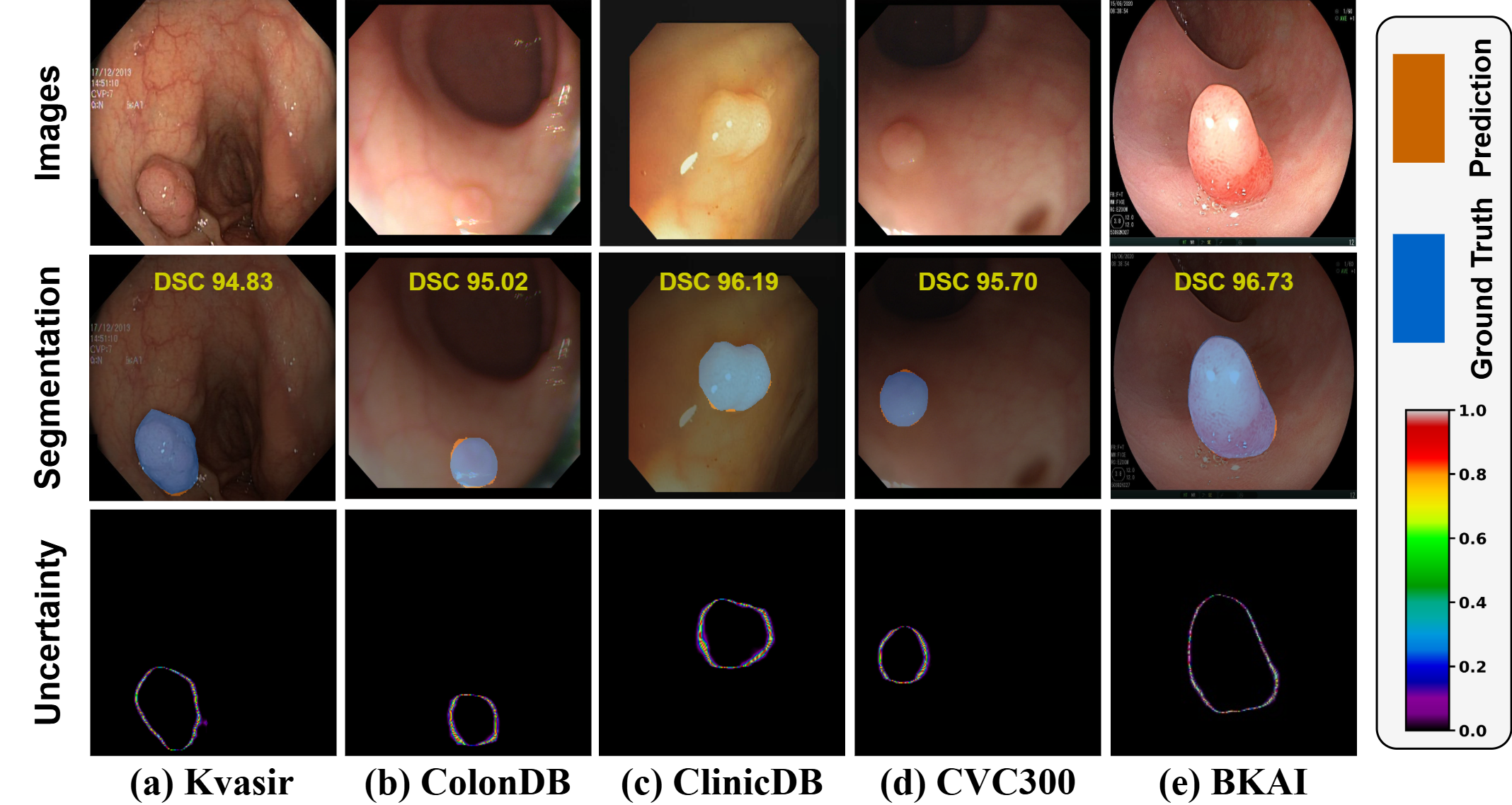}
    \caption{\textbf{Polyp Endoscopy Segmentation and uncertainty visualizations.} Uncertainty peaks along lesion boundaries and remains consistent across polyp endoscopy datasets, indicating reliable calibration and generalization.}
    \label{fig:segexamples-endoscopy}
\end{figure*}

\section{Effect of Different Confidence-Weighted Attention Mechanisms}
Table~\ref{tab:conf-attn} examines the impact of incorporating uncertainty into the attention computation in different manners. Our difference-based formulation yields the best performance across in-distribution (ID), out-of-distribution (OOD), and harmonic mean (HM) DSC scores. This approach adjusts attention weights by directly penalizing high-variance (low-confidence) regions, encouraging the model to focus on more reliable feature correspondences. For comparison, the \emph{weight scaling} variant applies an uncertainty-dependent multiplicative attenuation to the attention matrix:
\[
A^{\text{scaled}}
= \texttt{softmax}\!\left(
    S_{\mu}
\right) \;\oslash\; \bigl(1 + \beta S_{\sigma}\bigr),
\]
where \(\oslash\) denotes element-wise division. This strategy offers slightly lower gains compared to the proposed difference-based approach, while omitting uncertainty entirely reduces robustness under domain shifts. These results highlight that explicitly encoding confidence into attention promotes more stable and trustworthy segmentation performance.

\input{tables/conf_attn}

\section{Error vs. Uncertainty Correlation}

\begin{figure}[h]
    \centering
    \includegraphics[width=0.98\linewidth]{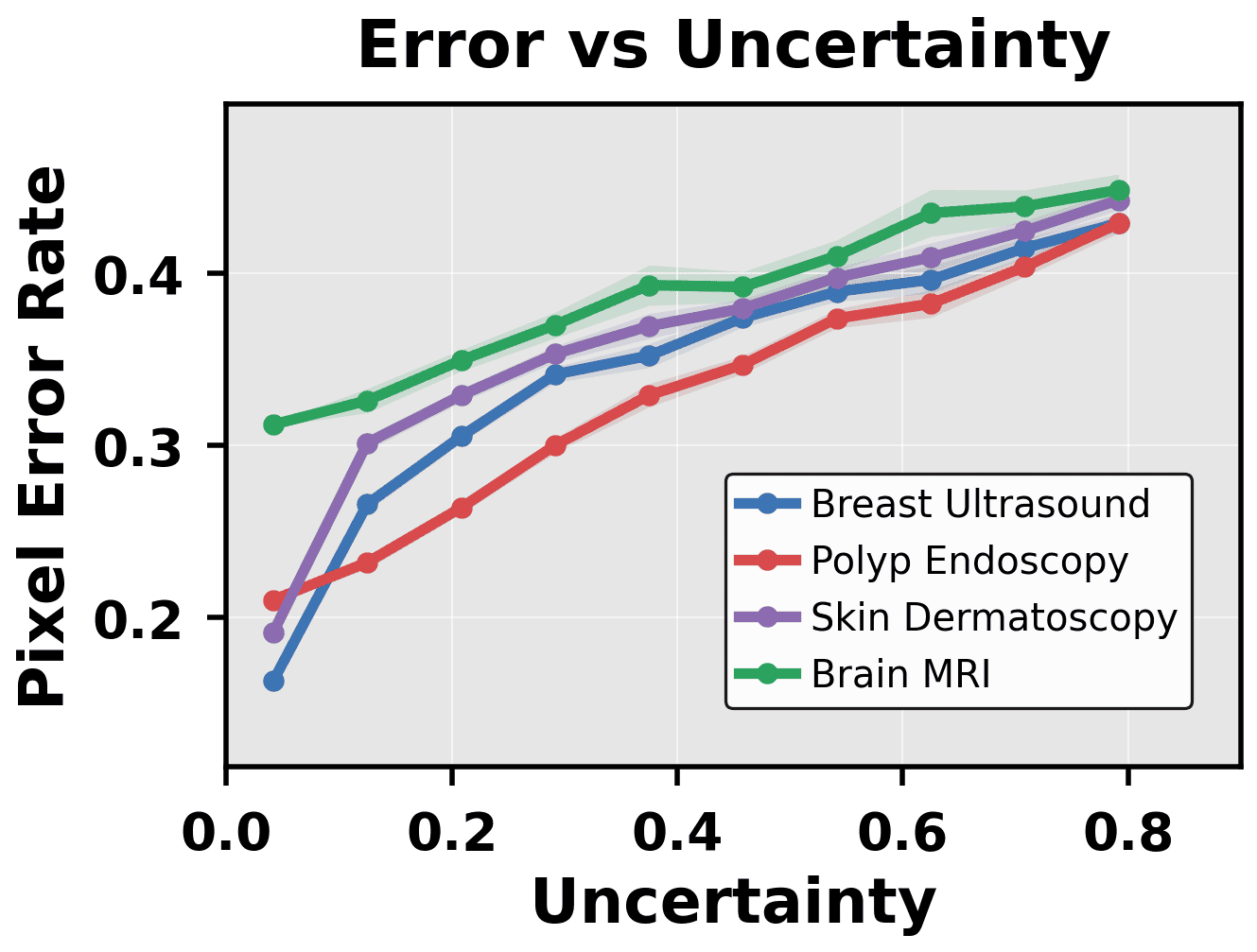}
    \caption{
        Relationship between predictive uncertainty and pixel-level segmentation error across imaging domains.}
    \label{fig:uncertainty-error-corr}
\end{figure}

\noindent
The model’s uncertainty maps exhibit a strong correlation with segmentation errors, as shown in Fig.~\ref{fig:uncertainty-error-corr}. Across all ID and OOD datasets, the Pearson correlations between uncertainty and error are consistently high: 0.9248 (Breast Ultrasound), 0.9921 (Polyp Endoscopy), 0.9201 (Skin Dermatoscopy), and 0.9885 (Brain MRI), all with $p$ $<$ 0.001. These strong correlations indicate that regions of elevated uncertainty reliably align with areas of higher prediction error, confirming that the uncertainty estimates meaningfully reflect model confidence and can support downstream tasks such as boundary refinement, error correction, and active learning.

\section{Deterministic vs. Probabilistic Confidence}
As shown in Fig. \ref{fig:conf-examples}, the probabilistic variant of \texttt{MedCLIPSeg} demonstrates superior handling of both overconfidence and underconfidence compared to its deterministic counterpart. Explicitly modeling predictive uncertainty suppresses spurious activations in non-lesion regions (reducing false positives) while recovering missed lesion boundaries (reducing false negatives). This leads to more balanced confidence calibration, smoother segmentation contours, and lower combined error rates across diverse ultrasound datasets.

\section{Additional Visualization Examples}
We further evaluate \texttt{MedCLIPSeg} under cross-domain conditions using polyp endoscopy and breast ultrasound datasets. As shown in Figs.~\ref{fig:segexamples-breastus} and \ref{fig:segexamples-endoscopy}, the model maintains strong segmentation quality despite noticeable domain shifts in texture, lighting, and instrument artifacts. Uncertainty maps remain concentrated along polyp and tumor boundaries, reflecting well-calibrated confidence and robust generalization to unseen endoscopic and ultrasound environments.

\section{Effect of Supervised Segmentation}
Table~\ref{tab:seg-sup} highlights the critical role of supervised segmentation annotations in medical image segmentation. When we remove the segmentation loss $\mathcal{L}_{\text{Seg}}$ and train the model solely with the soft patch-level contrastive objective $\mathcal{L}_{\text{SoftCon}}$, performance collapses to $\sim\!20\%$ DSC in-distribution and below $13\%$ out-of-distribution. Although purely contrastive or self-supervised objectives can be sufficient for natural-image segmentation, where object boundaries are often distinct and semantic categories are well separated and diverse, they are fundamentally insufficient in the medical domain. Medical boundaries are subtle, low-contrast, and frequently ambiguous, with fine-grained structures that require pixel-accurate supervision to disambiguate anatomy from imaging artifacts and surrounding tissues. Incorporating $\mathcal{L}_{\text{Seg}}$ provides this necessary spatial guidance, enabling the model to learn clinically meaningful decision boundaries and yielding dramatic gains of more than \textbf{+69\% DSC} ID and \textbf{+66\% DSC} OOD. These results clearly demonstrate that segmentation annotations remain indispensable for achieving reliable and generalizable medical image segmentation performance.
\input{tables/no_seg}

\input{tables/additional_baselines}
\section{Additional Baselines}
We include several additional baselines: a \underline{non-VLM} nnU-Net with checkpoint ensembling \cite{zhao2022efficient}, Ariadne's Thread \cite{zhong2023ariadne}, EviVLM \cite{evivlm}, VLSM-Ensemble \cite{dietlmeier2025vlsmensemble}, and our framework's variant with deterministic adapters and an \underline{evidential} mask head at the end for pixel-wise uncertainty estimation \underline{without Monte Carlo (MC) sampling}. We evaluate them all for robustness and uncertainty quality (\textbf{Spearman} correlation with errors) in Table \ref{tab:other-baselines}, and show that \texttt{MedCLIPSeg} consistently outperforms the others in both aspects.

\input{tables/additional_metrics}
\section{Additional Evaluation Metrics}
Table \ref{tab:other-metrics} reports \textbf{(Sensitivity, Specificity, F1)} for the top three baselines under domain generalization, supporting improved boundary localization by \texttt{MedCLIPSeg} under domain shifts.

\section{Sample Text Prompts}

Below, we provide one representative text prompt from each of the 16 datasets:
\\ \\
{
``\texttt{one small pink round polyp, located in right of the image
}"\\ \\
\noindent
``\texttt{A pituitary tumor is present in the center region of the brain.}"\\ \\
\noindent
``\texttt{Presence of a benign lesion located at the upper section.}"\\ \\
\noindent
``\texttt{Detected a malignant tumor positioned towards the center side.}"\\ \\
\noindent
``\texttt{One small rectangle-shaped regular tumor at the left in the breast ultrasound image.
}"\\ \\
\noindent
``\texttt{Findings indicate a benign tumor situated in the center area.}"\\ \\
\noindent
``\texttt{no irregularities detected on MRI scan}"\\ \\
\noindent
``\texttt{one small white triangular polyp, located in center of the image}"\\ \\
\noindent
``\texttt{one small pink circle polyp, located in center of the image}"\\ \\
\noindent
``\texttt{one small white circle polyp, located in center of the image}"\\ \\
\noindent
``\texttt{Bilateral pulmonary infection involving two regions with involvement of all left lung and all right lung}"\\ \\
\noindent
``\texttt{Endoscopic ultrasound showing heterogeneous mass in center}"\\ \\
\noindent
``\texttt{one medium red rectangular skin melanoma which is a spot with dark speckles located in top right of the image }"\\ \\
\noindent
``\texttt{one medium white round polyp, located in left of the image}"\\ \\
\noindent
``\texttt{Detected malignant lesion located at the center area.}"\\ \\
\noindent
``\texttt{Presence of a red skin melanoma positioned in the center part.}"\\ \\
\noindent
}

\section{Per-dataset Efficiency Results}
Tables~\ref{tab:full_results_efficiency_10}, \ref{tab:full_results_efficiency_25}, \ref{tab:full_results_efficiency_50}, and \ref{tab:full_results_efficiency_100} present detailed segmentation performance for each dataset across varying levels of labeled supervision (10\%, 25\%, 50\%, and 100\%). We report Dice Similarity Coefficient (DSC) and Normalized Surface Distance (NSD) scores to evaluate both volumetric overlap and boundary accuracy. This breakdown highlights the data-efficiency behavior of different model families, ranging from unimodal CNN and transformer baselines to text-driven and CLIP-based approaches. \texttt{MedCLIPSeg} consistently achieves the highest or second-highest performance across nearly all datasets and label fractions, demonstrating its robustness to annotation scarcity and strong cross-domain adaptability.
\input{tables/all_results_efficiency}

%% file: tables/dataset_details.tex
\begin{table*}[t]
    \centering
    \caption{\textbf{Summary of medical datasets:} Overview of datasets used in the data-efficiency, fully supervised, and domain generalization benchmarks.
    For \emph{data-efficiency} experiments, values in parentheses under \textit{Train} and \textit{Validation} indicate the number of samples corresponding to (10\%, 25\%, 50\%) of the full splits; other sections report full counts.}
    \label{tab:datasets}
    \tablestyle{-4pt}{1}
    \arrayrulecolor{black}
    \setlength\arrayrulewidth{1pt}
    \addtolength{\tabcolsep}{+12pt}
    \resizebox{0.98\linewidth}{!}{
    \begin{tabular}{l c c c c c}
    \toprule
    \textbf{Dataset} & \textbf{Train} & \textbf{Validation} & \textbf{Test} & \textbf{Modality} & \textbf{Organ} \\
    \midrule
    \rowcolor{gray!15} \multicolumn{6}{c}{\texttt{\textbf{Data-Efficiency Evaluation}}} \\
    BUSI \cite{busi}                & (62, 156, 312)         & (7, 19, 39)         & 78     & Ultrasound  & Breast \\
    BTMRI \cite{Cheng2017}          & (273, 684, 1,369)      & (132, 330, 660)     & 1,005  & MRI          & Brain \\
    ISIC \cite{isic1,isic2}         & (80, 202, 404)         & (9, 22, 45)         & 379    & Dermatoscopy & Skin \\
    Kvasir-SEG \cite{jha2019kvasir} & (80, 200, 400)         & (10, 25, 50)        & 100    & Endoscopy    & Colon \\
    QaTa-COV19 \cite{qatacov19}     & (571, 1,429, 2,858)    & (142, 357, 714)     & 2,113  & X-ray        & Chest \\
    EUS \cite{jaramillo2020endoscopic}
                                   & (2,631, 6,579, 13,159) & (175, 439, 879)     & 10,090 & Ultrasound   & Pancreas \\
    \midrule
    \rowcolor{gray!15} \multicolumn{6}{c}{\texttt{\textbf{Fully Supervised}}} \\
    BUSI \cite{busi}                & 624     & 78     & 78     & Ultrasound  & Breast \\
    BTMRI \cite{Cheng2017}          & 2,738   & 1,321  & 1,005  & MRI          & Brain \\
    ISIC \cite{isic1,isic2}         & 809     & 90     & 379    & Dermatoscopy & Skin \\
    Kvasir-SEG \cite{jha2019kvasir} & 800     & 100    & 100    & Endoscopy    & Colon \\
    QaTa-COV19 \cite{qatacov19}     & 5,716   & 1,429  & 2,113  & X-ray        & Chest \\
    EUS \cite{jaramillo2020endoscopic}
                                   & 26,318  & 1,758  & 10,090 & Ultrasound   & Pancreas \\
    \midrule
    \rowcolor{gray!15} \multicolumn{6}{c}{\texttt{\textbf{Domain Generalization}}} \\
    BUSUC \cite{busuc}              & 567     & 122    & 122    & Ultrasound  & Breast \\
    BUSBRA \cite{busbra}            & 1,311   & 282    & 282    & Ultrasound  & Breast \\
    BUID \cite{buid}                & 162     & 35     & 35     & Ultrasound  & Breast \\
    UDIAT \cite{udiat}              & 113     & 25     & 25     & Ultrasound  & Breast \\
    BRISC \cite{brisc}              & 4,000   & 1,000  & 1,000  & MRI          & Brain \\
    UWaterlooSkinCancer \cite{waterloo1,waterloo2}
                                   & 132     & 0      & 41     & Dermatoscopy & Skin \\
    CVC-ColonDB \cite{colondb}      & 20      & 0      & 360    & Endoscopy    & Colon \\
    CVC-ClinicDB \cite{clinicdb}    & 490     & 61     & 61     & Endoscopy    & Colon \\
    CVC-300 \cite{cvc300}           & 6       & 0      & 60     & Endoscopy    & Colon \\
    BKAI \cite{bkai}                & 799     & 100    & 100    & Endoscopy    & Colon \\
    \bottomrule
    \end{tabular}
    }
\end{table*}

%% file: tables/computational_cost.tex
\begin{table}[t]
    \caption{\textbf{Comparison of computational complexity between different methods.} 
    Models using text or multimodal supervision are marked with a \cmark\ in the \textit{``Text?''} column. }
    \centering
    \tablestyle{-8pt}{1.1}
    \addtolength{\tabcolsep}{+10pt}
    \resizebox{0.98\linewidth}{!}{%
    \begin{tabular}{lcccc}
    \toprule
    \textbf{Model} & \textbf{Text?} & \textbf{Params. (M)} & \textbf{FLOPs (G)} & \textbf{Inf. Time (s)} \\
    \midrule
    UNet \cite{ronneberger2015unet} & \xmark & 14.8 & 50.3  & 0.55 \\
    UNet++ \cite{zhou2018unet++} & \xmark & 74.5 & 94.6 & 0.81 \\
    DeepLabv3 \cite{chen2017rethinking} & \xmark & 57.6 & 38.4 & 1.16 \\
    AttnUNet \cite{oktay2018attention} & \xmark & 34.9 & 101.9 & 0.77 \\
    nnUNet \cite{isensee2021nnu} & \xmark & 19.1 & 412.7 & 1.55 \\
    Swin-UNet \cite{cao2022swin} & \xmark & 82.3 & 67.3 & 1.38 \\
    TransUNet \cite{chen2021transunet} & \xmark & 105 & 56.7 & 1.22 \\
    LViT \cite{li2023lvit} & \cmark & 29.7 & 54.1 & 1.74 \\
    Ariadne's Thread \cite{zhong2023ariadne} & \cmark & 44.0 & 49.8 & 2.39 \\
    CLIPSeg \cite{luddecke2022image} & \cmark & 1.1 & 66.8 & 1.35 \\
    DenseCLIP \cite{rao2022denseclip} & \cmark & 89.7 & 66.7  & 1.50 \\
    ZegCLIP \cite{zhou2023zegclip} & \cmark & 10.6 & 67.6 & 1.68 \\
    SAN \cite{xu2023side} & \cmark & 8.2 & 90.0 & 1.46 \\
    MaPLe \cite{khattak2023maple} & \cmark & 7.1 & 66.9 & 1.45 \\
    MaPLe \cite{khattak2023maple} + Decoder & \cmark & 8.2 & 67.3 & 1.75 \\
    VLSM-Adapter \cite{dhakal2024vlsm} & \cmark & 5.0 & 68.4 & 1.32 \\
    CausalCLIPSeg \cite{chen2024causalclipseg} & \cmark & 57.2 & 158.3 & 4.39 \\
    CAT-Seg \cite{cho2024cat} & \cmark & 34.8 & 69.7 & 2.34 \\
    \rowcolor{lightpurple}\texttt{MedCLIPSeg} (Ours) & \cmark & \textbf{18.7} & \textbf{73.6} & \textbf{1.51} \\
    \bottomrule
    \end{tabular}
    }
    \label{tab:computational-cost}
\end{table}

%% file: tables/text_prompt_generation_psuedocode.tex
\begin{algorithm}[H]
\caption{Text Prompt Generation}
\label{alg:text_prompt_generation}
\resizebox{0.98\linewidth}{!}{%
\begin{minipage}{0.98\linewidth}
\begin{algorithmic}[1]
\STATE \textbf{Inputs:} \texttt{Images}, \texttt{Masks} (optional), \texttt{Labels} (optional)
\STATE \textbf{Goal:} Produce one caption per image using LLM templates + simple attributes

\vspace{3pt}
\STATE \textbf{Step 1: Templates (once per dataset)}
\STATE Query an LLM to write a few short templates with placeholders:
\{class\}, \{location\}, \{number\}, \{shape\}, \{color\}.
\STATE Provide separate ``normal'' and ``lesion'' templates.

\vspace{3pt}
\STATE \textbf{Step 2: Attribute extraction (per image)}
\FOR{each image in \texttt{Images}}
  \IF{a corresponding mask exists}
    \STATE \textbf{Class}: use label if available; otherwise ``lesion'' if uniform.
    \STATE \textbf{Location}: coarse region from mask (e.g., upper/lower/left/right/center).
    \STATE \textbf{Number}: count connected components (single/multiple).
    \STATE \textbf{Shape}: coarse shape cue (e.g., round/irregular).
    \STATE \textbf{Color}: overall brightness/tone relative to background.
  \ELSE
    \STATE Mark as ``normal'' (no lesion mask).
  \ENDIF
\ENDFOR

\vspace{3pt}
\STATE \textbf{Step 3: Fill templates (per image)}
\FOR{each image}
  \IF{normal}
    \STATE Choose a ``normal'' template; save caption.
  \ELSE
    \STATE Choose a ``lesion'' template; replace placeholders with extracted attributes; save caption.
  \ENDIF
\ENDFOR

\vspace{3pt}
\STATE \textbf{Output:} A list of paired \texttt{(image, text, mask)} samples ready for training or evaluation..
\end{algorithmic}
\end{minipage}%
}
\end{algorithm}

%% file: tables/prompt_design_overview.tex
\begin{table*}
\centering
\tablestyle{-6pt}{1.2} % slightly increased for uniform padding
\addtolength{\tabcolsep}{+10pt}
\caption{\textbf{Prompt design taxonomy with examples.} Each configuration illustrates how wording choices (conciseness, spatial detail, contradictions, and noise) affect the semantics supplied to the model.}
\resizebox{0.98\linewidth}{!}{%
\renewcommand{\arraystretch}{1.25} % ensures equal vertical space between rows
\begin{NiceTabular}{l p{0.22\linewidth} p{0.28\linewidth} p{0.28\linewidth}}
\toprule
\textbf{Design Type} & \textbf{Description Style} & \textbf{Example (Normal)} & \textbf{Example (Tumor)} \\
\midrule
\texttt{Original} & Balanced, accurate &
``\texttt{The breast appears normal with no signs of lesions.}'' &
``\texttt{A malignant tumor is present in the upper-left region of the breast.}'' \\

\texttt{Underdescriptive} & Minimal, label-only &
``\texttt{Normal breast.}'' &
``\texttt{Tumor present.}'' \\

\texttt{Overdescriptive} & Verbose, redundant &
``\texttt{The breast tissue appears entirely healthy, with homogeneous echotexture throughout.}'' &
``\texttt{A clearly defined malignant tumor with irregular boundaries located in the upper-left quadrant.}'' \\

\texttt{Contradictory} & Incorrect/Conflicting info &
``\texttt{Normal breast tissue with a visible lesion in the image.}'' &
``\texttt{Malignant tumor detected, but breast appears completely normal.}'' \\

\texttt{Missing Location} & No spatial info &
``\texttt{The breast appears normal with no signs of lesions.}'' &
``\texttt{A malignant tumor is detected in the breast.}'' \\
\bottomrule
\end{NiceTabular}
}
\label{tab:prompt-design-examples}
\end{table*}

%% file: tables/sampling_cost.tex
\begin{table}[htpb]
\centering
\caption{
\textbf{Performance-cost tradeoff under MC sampling}
}
\tablestyle{-12pt}{1.1}
\addtolength{\tabcolsep}{+14pt}
\resizebox{0.98\linewidth}{!}{%
\begin{NiceTabular}{c|cccc|c}
\toprule
 \textbf{Samples}                   & \textbf{Runtime (s/batch)}            & \textbf{FPS} & \textbf{HM DSC (\%)} & \textbf{HM Spearman (\%)} & \textcolor{teal}{\textbf{3D Runtime (s/vol)}}  \\ \midrule
5 & 1.78 & 24.92 & 83.52 & 83.07 & \textcolor{teal}{0.98} \\
10 & 2.20 & 14.35 & 83.66 & 83.44 & \textcolor{teal}{1.03} \\
20 & 4.08 & 7.64 & 83.71 &  83.52 & \textcolor{teal}{1.97} \\
30 & 6.01 & 5.23 & 83.76 & 83.84 & \textcolor{teal}{2.90} \\
\bottomrule
\end{NiceTabular}%
}
\label{tab:resource_sampling}
\end{table}

%% file: tables/gating_init.tex
\begin{table}[htpb]
\centering
\caption{
\textbf{Effect of the gating initialization}
}
\tablestyle{-12pt}{1.1}
\addtolength{\tabcolsep}{+14pt}
\resizebox{0.98\linewidth}{!}{%
\begin{NiceTabular}{c|ccc}
\toprule
 \textbf{Gating Init. ($g$)}                   & \textbf{ID DSC (\%)}           & \textbf{OOD DSC (\%)}          & \textbf{HM DSC (\%)}    \\ \midrule
 \texttt{sigmoid(-0.5)} & 88.79         & 74.65          &   81.11        \\
\rowcolor{lightpurple} \texttt{sigmoid(0)} & \textbf{89.11}          & \textbf{79.02}          &  \textbf{83.76}         \\
\texttt{sigmoid(0.5)} &  88.93         &  77.51         &     82.83      \\
\bottomrule
\end{NiceTabular}%
}
\label{tab:gating-init}
\end{table}

%% file: tables/twoway.tex
\begin{table}[htpb]
\centering
\caption{
\textbf{Effect of the two-way attention mechanism}
}
\tablestyle{-12pt}{1.1}
\addtolength{\tabcolsep}{+14pt}
\resizebox{0.98\linewidth}{!}{%
\begin{NiceTabular}{c|ccc}
\toprule
 \textbf{Two-way Mechanism}                   & \textbf{ID DSC (\%)}           & \textbf{OOD DSC (\%)}          & \textbf{HM DSC (\%)}    \\ \midrule
 \texttt{None} & 88.71         & 77.71         &   82.85       \\
  \texttt{Text First} & 88.55         & 76.99          &   82.37        \\
\rowcolor{lightpurple} \texttt{Vision First} & \textbf{89.11}          & \textbf{79.02}          &  \textbf{83.76}         \\
\bottomrule
\end{NiceTabular}%
}
\label{tab:two-way}
\end{table}

%% file: tables/contrastive_loss.tex
\begin{table}[htpb]
\centering
\caption{
\textbf{Effect of the pooling strategies}
}
\tablestyle{-12pt}{1.1}
\addtolength{\tabcolsep}{+14pt}
\resizebox{0.98\linewidth}{!}{%
\begin{NiceTabular}{c|ccc}
\toprule
 \textbf{Contrastive Pooling}                   & \textbf{ID DSC (\%)}           & \textbf{OOD DSC (\%)}          & \textbf{HM DSC (\%)}    \\ \midrule
 \texttt{[CLS]} & 88.89         & 78.28         &   83.25       \\
  \texttt{Attention Pooling} & 88.73         & 75.60          &   81.64        \\
\rowcolor{lightpurple} \texttt{Average Pooling} & \textbf{89.11}          & \textbf{79.02}          &  \textbf{83.76}         \\
\bottomrule
\end{NiceTabular}%
}
\label{tab:con-loss}
\end{table}

%% file: tables/upscaling_blocks.tex
\begin{table}[htpb]
\centering
\caption{
\textbf{Effect of the number of upscaling layers}
}
\tablestyle{-12pt}{1.1}
\addtolength{\tabcolsep}{+14pt}
\resizebox{0.98\linewidth}{!}{%
\begin{NiceTabular}{c|ccc}
\toprule
 \textbf{Num. Upscale}                   & \textbf{ID DSC (\%)}           & \textbf{OOD DSC (\%)}          & \textbf{HM DSC (\%)}    \\ \midrule
 1 & 88.73         & 75.74        &   81.72       \\
  \rowcolor{lightpurple} 2 & \textbf{89.11}          & \textbf{79.02}          &  \textbf{83.76} \\
 3 & 88.64          & 74.99         &  81.24         \\
\bottomrule
\end{NiceTabular}%
}
\label{tab:upscale-blocks}
\end{table}

%% file: tables/adapter_dim.tex
\begin{table}[htpb]
\centering
\caption{
\textbf{Effect of the shared dimension in the \textit{PVL} adapters}
}
\tablestyle{-12pt}{1.1}
\addtolength{\tabcolsep}{+14pt}
\resizebox{0.98\linewidth}{!}{%
\begin{NiceTabular}{c|ccc}
\toprule
 \textbf{Adapter Dim. ($D_s$)}                   & \textbf{ID DSC (\%)}           & \textbf{OOD DSC (\%)}          & \textbf{HM DSC (\%)}    \\ \midrule
 64 & 87.76         & 74.63        &   80.66       \\
 128 & 88.68         & 76.44        &   82.11       \\
 192 & 88.93         & 76.01        &   81.96      \\
  \rowcolor{lightpurple} 256 & \textbf{89.11}          & \textbf{79.02}          &  \textbf{83.76} \\
 512 & 88.56          & 77.96         &  82.92         \\
\bottomrule
\end{NiceTabular}%
}
\label{tab:adapter-dim}
\end{table}

%% file: tables/conf_attn.tex
\begin{table}[htpb]
\centering
\caption{
\textbf{Effect of confidence-weighted attention mechanism}
}
\tablestyle{-12pt}{1.1}
\addtolength{\tabcolsep}{+14pt}
\resizebox{0.98\linewidth}{!}{%
\begin{NiceTabular}{c|ccc}
\toprule
 \textbf{Mechanism}                   & \textbf{ID DSC (\%)}           & \textbf{OOD DSC (\%)} &  \textbf{HM DSC (\%)} \\ \midrule
 \texttt{None} & 88.97 & 77.49 & 82.83 \\
 \texttt{Scaling} & 88.92         & 78.78          &  83.54        \\
\rowcolor{lightpurple} \texttt{Difference} & \textbf{89.11}          & \textbf{79.02}          &  \textbf{83.76}         \\
\bottomrule
\end{NiceTabular}%
}
\label{tab:conf-attn}
\end{table}

%% file: tables/no_seg.tex
\begin{table}[htbp]
\centering
\caption{
\textbf{Effect of the segmentation annotations}
}
\tablestyle{-12pt}{1.1}
\addtolength{\tabcolsep}{+14pt}
\resizebox{0.98\linewidth}{!}{%
\begin{NiceTabular}{c|ccc}
\toprule
 \textbf{$\mathcal{L}_{Seg}$?}                   & \textbf{ID DSC (\%)}           & \textbf{OOD DSC (\%)}          & \textbf{HM DSC (\%)}    \\ \midrule
 \xmark & 19.84         & 12.68          &   15.47        \\
\rowcolor{lightpurple} \cmark & \textbf{89.11}          & \textbf{79.02}          &  \textbf{83.76}         \\
\bottomrule
\end{NiceTabular}%
}
\label{tab:seg-sup}
\end{table}

%% file: tables/additional_baselines.tex
\begin{table}[htpb]
\centering
\caption{
\textbf{Domain generalization with more baselines (\%)}
}
\tablestyle{-12pt}{1.1}
\addtolength{\tabcolsep}{+14pt}
\resizebox{0.98\linewidth}{!}{%
\begin{NiceTabular}{ccccccc}
\toprule
\multirow{2}{*}{\textbf{Method}} 
& \multicolumn{2}{c}{\textbf{ID}} 
& \multicolumn{2}{c}{\textbf{OOD}} 
& \multicolumn{2}{c}{\textbf{HM}} \\
\cmidrule(lr){2-3}
\cmidrule(lr){4-5}
\cmidrule(lr){6-7}
& \textbf{DSC} & \textbf{Spearman}
& \textbf{DSC} & \textbf{Spearman}
& \textbf{DSC} & \textbf{Spearman} \\
\midrule
% Ariadne's Thread & 57.26 & 20.67 & 49.36 & 22.29 & 22.19 & 77.42 & 23.08 & 22.99 & 20.18 & 22.19 & 69.96 & 25.89 & 68.37 & 43.43 \\
Ariadne's Thread \cite{zhong2023ariadne}
& 68.25 & --
& 27.24 & --
& 38.94 & -- \\
EviVLM \cite{evivlm}
& 84.06 & -- 
& 54.47 & -- 
& 66.10 & -- \\
VLSM-Ensemble \cite{dietlmeier2025vlsmensemble}
& 87.36 & -- 
& 63.24 & -- 
& 73.37 & -- \\
nnUNet + Ensembling \cite{zhao2022efficient}
& 86.50 & 66.74 
& 74.20 & 55.22 
& 79.80 & 60.44 \\
MedCLIPSeg (Evidential) 
& 88.18 & 78.49 
& 76.61 & 76.79 
& 81.99 & 77.63 \\
\rowcolor{lightpurple}
\texttt{MedCLIPSeg} (Ours) 
& \textbf{89.11} & \textbf{87.57}
& \textbf{79.02} & \textbf{80.41}
& \textbf{83.76} & \textbf{83.84} \\
\bottomrule
\end{NiceTabular}%
}
\label{tab:other-baselines}
\end{table}

%% file: tables/additional_metrics.tex
\begin{table}[htpb]
\centering
\caption{
\textbf{Domain generalization with different metrics (\%)}
}
\tablestyle{-12pt}{1.1}
\addtolength{\tabcolsep}{+14pt}
\resizebox{0.98\linewidth}{!}{%
\begin{NiceTabular}{c|ccc}
\toprule
 \textbf{Method}                   & \textbf{ID}           & \textbf{OOD}          & \textbf{HM}    \\ \midrule

SAN [\textcolor{teal}{68}]
& (86.9, 88.9, 84.5) 
& (74.3, 83.9, 69.9) 
& (80.1, 86.3, 76.5) \\

VLSM-Adapter [\textcolor{teal}{18}] 
& (88.5, 87.9, 85.8) 
& (80.3, 80.8, 73.3) 
& (84.2, 84.2, 79.0) \\

CAT-Seg [\textcolor{teal}{15}] 
& (87.2, 89.1, 86.1) 
& (81.1, 82.6, 74.6)
& (84.0, 85.7, 79.9) \\

\rowcolor{lightpurple}
\texttt{MedCLIPSeg} (Ours) 
& (89.9, 90.8, 89.1) 
& (86.2, 80.7, 79.0) 
& (88.0, 85.5, 83.8) \\

\bottomrule
\end{NiceTabular}%
}
\label{tab:other-metrics}
\end{table}

%% file: tables/all_results_efficiency.tex
\begin{table*}[t!]
\caption{\textbf{Per-dataset segmentation with 10\% Labeled Data:} 
This table reports DSC and NSD values (\%) across six medical image segmentation benchmarks. 
All baseline methods are trained using 10\% of the ground-truth annotations. Best results are in \textbf{bold}, and second-best are \underline{underlined}.}
\centering
\tablestyle{-27pt}{1.1}
\addtolength{\tabcolsep}{+30pt}
\renewcommand{\arraystretch}{1.5}
\resizebox{0.98\linewidth}{!}{%
\begin{tabular}{ccccccccccccc}
\toprule
\textbf{Method} & \multicolumn{2}{c}{\textbf{BUSI}} & \multicolumn{2}{c}{\textbf{BTMRI}}  & \multicolumn{2}{c}{\textbf{ISIC}} & \multicolumn{2}{c}{\textbf{Kvasir-SEG}} & \multicolumn{2}{c}{\textbf{QaTa-COV19}} & \multicolumn{2}{c}{\textbf{EUS}} \\ 
\cmidrule(lr){2-3} \cmidrule(lr){4-5} \cmidrule(lr){6-7} \cmidrule(lr){8-9} \cmidrule(lr){10-11} \cmidrule(lr){12-13}
& \textbf{DSC} $\uparrow$ & \textbf{NSD} $\uparrow$ & \textbf{DSC} $\uparrow$ & \textbf{NSD} $\uparrow$ & \textbf{DSC} $\uparrow$ & \textbf{NSD} $\uparrow$ & \textbf{DSC} $\uparrow$ & \textbf{NSD} $\uparrow$ & \textbf{DSC} $\uparrow$ & \textbf{NSD} $\uparrow$ 
& \textbf{DSC} $\uparrow$ & \textbf{NSD} $\uparrow$ \\ \midrule
\rowcolor{gray!15} \multicolumn{13}{c}{\texttt{\textbf{Unimodal Approaches}}} \\
UNet \cite{ronneberger2015unet}         & 49.33 & 52.13 & 64.49 & 69.12 & 79.43 & 81.93 & 40.66 & 44.16 & 71.41 & 77.18 & 60.40 & 62.07 \\
UNet++ \cite{zhou2018unet++}       & 53.80 & 57.58 & 62.23 & 66.11 & 82.83 & 85.54 & 46.27 & 49.28 & 69.22 & 74.99 & 67.96 & 69.00 \\
DeepLabv3 \cite{chen2017rethinking}    & 42.45 & 45.64 & 61.96 & 66.47 & 84.62 & 87.72 & 45.14 & 48.32 & 68.51 & 74.38 & 65.22 & 66.50 \\
AttnUNet \cite{oktay2018attention}     & 54.66 & 58.17 & 58.68 & 62.77 & 85.16 & 88.08 & 42.46 & 45.68 & 70.86 & 76.37 & 64.85 & 66.44 \\
nnUNet \cite{isensee2021nnu} & 56.32 & 60.78 & 81.44 & 86.38 & 88.67 & 91.61 & 74.15 & 78.48 & 70.20 & 75.81 & 69.94 & 71.14 \\
Swin-UNet \cite{cao2022swin}  & 39.87 & 45.20 & 41.26 & 45.83 & 81.04 & 84.12 & 36.84 & 43.05 & 62.10 & 70.43 & 57.14 & 58.81 \\
TransUNet \cite{chen2021transunet} & 39.61 & 43.19 & 55.04 & 58.65 & 84.43 & 87.30 & 47.48 & 51.56 & 54.50 & 61.29 & 35.09 & 36.29 \\ \midrule
\rowcolor{gray!15}
\multicolumn{13}{c}{\texttt{\textbf{Generic Text-driven Approaches}}} \\
% LAVT \cite{yang2022lavt} & 55.50 & 57.72 & 69.03 & 71.88 & 86.17 & 87.23 & 49.00 & 50.50 & 70.95 & 77.19 & 81.28 & 81.97 \\
LViT \cite{li2023lvit} & 63.37 & 65.97 & 52.48 & 54.80 & 74.53 & 75.48 & 51.60 & 53.10 & 76.52 & 82.23 & 80.57 & 81.21 \\ 
Ariadne's Thread \cite{zhong2023ariadne}  & 35.51 & 36.39 & 58.70 & 60.01 & 66.28 & 67.25 & 74.98 & 76.40 & 59.86 & 63.13 & 76.12 & 76.68 \\ \midrule
\rowcolor{gray!15} \multicolumn{13}{c}{\texttt{\textbf{CLIP-Based Approaches}}} \\ 
CLIPSeg \cite{luddecke2022image} & 65.65 & 68.40 & 72.97 & 76.42 & 85.18 & 86.26 & 67.81 & 70.53 & 74.47 & 82.16 & 81.90 & 82.75 \\
DenseCLIP \cite{rao2022denseclip} & 55.09 & 57.41 & 60.56 & 61.87 & 88.54 & 89.56 & 73.73 & 75.80 & 66.95 & 73.86 & 62.18 & 63.48 \\
ZegCLIP \cite{zhou2023zegclip} & 46.20 & 48.13 & 70.74 & 73.46 & 79.16 & 80.07 & 68.37 & 70.29 & 69.19 & 75.88 & 33.84 & 34.49 \\
SAN \cite{xu2023side} & 66.99 & 69.56 & 77.92 & 81.76 & 89.20 & 90.21 & 66.79 & 69.36 & 72.36 & 78.78 & 71.51 & 72.17 \\
MaPLe \cite{khattak2023maple} & 55.70 & 57.98 & 70.14 & 71.57 & 86.50 & 87.50 & 59.82 & 62.05 & 67.11 & 74.24 & 58.37 & 59.16 \\
MaPLe \cite{khattak2023maple} + Decoder & 60.50 & 63.49 & 73.18 & 76.83 & 84.47 & 85.57 & 66.67 & 69.28 & 76.95 & 84.27 & 87.11 & 87.97 \\
VLSM-Adapter \cite{dhakal2024vlsm} & 63.85 & 66.60 & 73.19 & 76.81 & 86.81 & 87.95 & 71.74 & 74.44 & 74.90 & 82.06 & 76.31 & 77.14 \\
CausalCLIPSeg \cite{chen2024causalclipseg} & 51.29 & 53.02 & 73.97 & 77.27 & 84.89 & 85.86 & 60.35 & 62.24 & 70.58 & 77.18 & 82.61 & 84.17 \\ 
CAT-Seg \cite{cho2024cat} & 68.01 & 70.66 & 77.15 & 80.38 & 87.95 & 89.01 & 75.43 & 77.60 & 76.19 & 82.71 & 87.82 & 88.65 \\ \midrule
\rowcolor{lightpurple}
\texttt{MedCLIPSeg} (Ours) & 68.66 & 71.35 & 79.07 & 82.71 & 90.35 & 91.40 & 77.21 & 79.53 & 79.73 & 86.27 & 91.59 & 92.38 \\
\bottomrule
\end{tabular}
}
\label{tab:full_results_efficiency_10}
\end{table*}

\begin{table*}[t!]
\caption{\textbf{Per-dataset segmentation with 25\% Labeled Data:} 
This table reports DSC and NSD values (\%) across six medical image segmentation benchmarks. 
All baseline methods are trained using 25\% of the ground-truth annotations. Best results are in \textbf{bold}, and second-best are \underline{underlined}.}
\centering
\tablestyle{-27pt}{1.1}
\addtolength{\tabcolsep}{+30pt}
\renewcommand{\arraystretch}{1.5}
\resizebox{0.98\linewidth}{!}{%
\begin{tabular}{ccccccccccccc}
\toprule
\textbf{Method} & \multicolumn{2}{c}{\textbf{BUSI}} & \multicolumn{2}{c}{\textbf{BTMRI}}  & \multicolumn{2}{c}{\textbf{ISIC}} & \multicolumn{2}{c}{\textbf{Kvasir-SEG}} & \multicolumn{2}{c}{\textbf{QaTa-COV19}} & \multicolumn{2}{c}{\textbf{EUS}} \\ 
\cmidrule(lr){2-3} \cmidrule(lr){4-5} \cmidrule(lr){6-7} \cmidrule(lr){8-9} \cmidrule(lr){10-11} \cmidrule(lr){12-13}
& \textbf{DSC} $\uparrow$ & \textbf{NSD} $\uparrow$ & \textbf{DSC} $\uparrow$ & \textbf{NSD} $\uparrow$ & \textbf{DSC} $\uparrow$ & \textbf{NSD} $\uparrow$ & \textbf{DSC} $\uparrow$ & \textbf{NSD} $\uparrow$ & \textbf{DSC} $\uparrow$ & \textbf{NSD} $\uparrow$ 
& \textbf{DSC} $\uparrow$ & \textbf{NSD} $\uparrow$ \\ \midrule
\rowcolor{gray!15} \multicolumn{13}{c}{\texttt{\textbf{Unimodal Approaches}}} \\
UNet \cite{ronneberger2015unet}  & 56.28 & 59.90 & 70.40 & 74.98 & 82.44 & 85.68 & 41.88 & 44.93 & 68.23 & 73.05 & 57.23 & 58.45 \\
UNet++ \cite{zhou2018unet++}  & 56.68 & 60.36 & 76.46 & 81.07 & 84.35 & 87.05 & 62.23 & 65.56 & 43.36 & 48.05 & 72.07 & 73.15 \\
DeepLabv3 \cite{chen2017rethinking} & 55.03 & 58.85 & 73.16 & 78.77 & 87.01 & 90.08 & 54.34 & 58.04 & 66.98 & 72.75 & 55.79 & 56.10 \\
AttnUNet \cite{oktay2018attention}  & 62.55 & 66.75 & 64.24 & 68.11 & 85.72 & 88.59 & 55.17 & 58.94 & 55.00 & 60.14 & 67.16 & 68.65 \\
nnUNet \cite{isensee2021nnu} & 62.28 & 66.29 & 83.20 & 89.12 & 89.85 & 92.76 & 80.96 & 84.91 & 72.84 & 78.41 & 71.26 & 72.44 \\
Swin-UNet \cite{cao2022swin}  & 37.56 & 42.24 & 66.20 & 71.85 & 80.35 & 83.33 & 42.49 & 47.95 & 53.94 & 60.36 & 47.59 & 49.71 \\
TransUNet \cite{chen2021transunet} & 46.21 & 49.91 & 58.33 & 62.24 & 86.04 & 88.82 & 51.51 & 55.95 & 50.09 & 56.19 & 39.33 & 40.57 \\ \midrule
\rowcolor{gray!15}
\multicolumn{13}{c}{\texttt{\textbf{Generic Text-driven Approaches}}} \\
% LAVT \cite{yang2022lavt} & 68.44 & 71.00 & 72.29 & 75.50 & 87.28 & 88.32 & 47.35 & 48.78 & 76.47 & 82.73 & 82.89 & 83.60 \\
LViT \cite{li2023lvit} & 62.31 & 64.64 & 76.21 & 79.53 & 81.02 & 81.97 & 72.52 & 74.43 & 78.99 & 84.55 & 82.94 & 83.60 \\
Ariadne's Thread \cite{zhong2023ariadne} & 39.01 & 40.04 & 58.70 & 60.01 & 68.53 & 69.45 & 75.71 & 76.11 & 60.06 & 64.29 & 76.54 & 77.14 \\ \midrule
\rowcolor{gray!15} \multicolumn{13}{c}{\texttt{\textbf{CLIP-Based Approaches}}} \\ 
CLIPSeg \cite{luddecke2022image} & 70.35 & 73.18 & 74.91 & 78.40 & 86.45 & 87.59 & 73.67 & 76.41 & 78.77 & 85.74 & 85.72 & 86.72 \\
DenseCLIP \cite{rao2022denseclip} & 58.27 & 59.94 & 64.97 & 67.32 & 88.98 & 89.69 & 78.26 & 80.40 & 65.92 & 72.66 & 64.96 & 66.19 \\
ZegCLIP \cite{zhou2023zegclip} & 49.86 & 51.88 & 73.18 & 76.11 & 79.39 & 80.24 & 71.95 & 73.75 & 72.99 & 80.09 & 87.37 & 88.00 \\
SAN \cite{xu2023side} & 64.43 & 67.09 & 81.15 & 84.96 & 90.65 & 91.69 & 77.48 & 79.76 & 74.35 & 80.57 & 68.73 & 69.40 \\
MaPLe \cite{khattak2023maple} & 63.94 & 66.42 & 72.87 & 73.91 & 87.89 & 88.89 & 71.68 & 73.85 & 67.43 & 74.68 & 65.39 & 65.93 \\
MaPLe \cite{khattak2023maple} + Decoder & 67.12 & 69.75 & 79.78 & 83.74 & 87.89 & 88.98 & 74.96 & 77.57 & 79.52 & 86.15 & 88.58 & 89.39 \\ 
VLSM-Adapter \cite{dhakal2024vlsm} & 63.48 & 66.17 & 79.50 & 83.33 & 89.44 & 90.52 & 76.45 & 78.93 & 78.14 & 84.67 & 78.74 & 79.54 \\
CausalCLIPSeg \cite{chen2024causalclipseg} & 57.76 & 59.67 & 76.15 & 79.57 & 85.98 & 86.93 & 68.57 & 70.35 & 76.17 & 82.95 & 83.86 & 84.54 \\  
CAT-Seg \cite{cho2024cat} & 73.08 & 75.76 & 80.07 & 83.59 & 89.61 & 90.67 & 80.11 & 82.50 & 79.12 & 85.45 & 84.73 & 85.54 \\ \midrule
\rowcolor{lightpurple}
\texttt{MedCLIPSeg} (Ours) & 77.73 & 80.29 & 83.93 & 87.69 & 91.00 & 92.04 & 84.21 & 86.46 & 81.83 & 88.01 & 91.79 & 92.61 \\
\bottomrule
\end{tabular}
}
\label{tab:full_results_efficiency_25}
\end{table*}

\begin{table*}[t!]
\caption{\textbf{Per-dataset segmentation with 50\% Labeled Data:} 
This table reports DSC and NSD values (\%) across six medical image segmentation benchmarks. 
All baseline methods are trained using 50\% of the ground-truth annotations. Best results are in \textbf{bold}, and second-best are \underline{underlined}.}
\centering
\tablestyle{-27pt}{1.1}
\addtolength{\tabcolsep}{+30pt}
\renewcommand{\arraystretch}{1.5}
\resizebox{0.98\linewidth}{!}{%
\begin{tabular}{ccccccccccccc}
\toprule
\textbf{Method} & \multicolumn{2}{c}{\textbf{BUSI}} & \multicolumn{2}{c}{\textbf{BTMRI}}  & \multicolumn{2}{c}{\textbf{ISIC}} & \multicolumn{2}{c}{\textbf{Kvasir-SEG}} & \multicolumn{2}{c}{\textbf{QaTa-COV19}} & \multicolumn{2}{c}{\textbf{EUS}} \\ 
\cmidrule(lr){2-3} \cmidrule(lr){4-5} \cmidrule(lr){6-7} \cmidrule(lr){8-9} \cmidrule(lr){10-11} \cmidrule(lr){12-13}
& \textbf{DSC} $\uparrow$ & \textbf{NSD} $\uparrow$ & \textbf{DSC} $\uparrow$ & \textbf{NSD} $\uparrow$ & \textbf{DSC} $\uparrow$ & \textbf{NSD} $\uparrow$ & \textbf{DSC} $\uparrow$ & \textbf{NSD} $\uparrow$ & \textbf{DSC} $\uparrow$ & \textbf{NSD} $\uparrow$ 
& \textbf{DSC} $\uparrow$ & \textbf{NSD} $\uparrow$ \\ \midrule
\rowcolor{gray!15} \multicolumn{13}{c}{\texttt{\textbf{Unimodal Approaches}}} \\
UNet \cite{ronneberger2015unet}   & 60.46 & 64.38 & 81.47 & 86.06 & 86.85 & 89.93 & 72.12 & 75.51 & 66.36 & 71.15 & 62.38 & 63.80 \\
UNet++ \cite{zhou2018unet++}  & 63.63 & 67.16 & 80.21 & 84.61 & 88.29 & 91.23 & 74.53 & 77.72 & 63.80 & 67.71 & 68.42 & 69.43 \\
DeepLabv3 \cite{chen2017rethinking} & 55.83 & 60.21 & 77.82 & 83.56 & 86.14 & 89.06 & 67.72 & 71.90 & 64.86 & 70.29 & 59.12 & 60.39 \\
AttnUNet \cite{oktay2018attention}  & 59.81 & 63.46 & 72.96 & 77.24 & 87.28 & 90.48 & 74.79 & 78.66 & 64.71 & 69.95 & 68.52 & 69.99 \\
nnUNet \cite{isensee2021nnu}  & 68.15 & 71.97 & 85.30 & 91.18 & 90.38 & 93.21 & 83.45 & 87.30 & 74.44 & 79.89 & 71.42 & 72.50 \\
Swin-UNet \cite{cao2022swin} & 41.86 & 48.11 & 57.76 & 64.96 & 85.52 & 89.14 & 50.04 & 55.25 & 53.67 & 61.58 & 46.51 & 48.44 \\
TransUNet \cite{chen2021transunet} & 41.95 & 45.95 & 62.60 & 67.63 & 86.43 & 89.27 & 54.78 & 59.41 & 52.98 & 58.79 & 32.58 & 34.73 \\ \midrule
\rowcolor{gray!15}
\multicolumn{13}{c}{\texttt{\textbf{Generic Text-driven Approaches}}} \\
% LAVT \cite{yang2022lavt} & 74.29 & 76.92 & 79.79 & 83.31 & 88.85 & 89.88 & 67.59 & 69.41 & 79.10 & 85.18 & 84.81 & 85.58 \\
LViT \cite{li2023lvit} & 62.74 & 65.29 & 78.89 & 82.17 & 89.18 & 90.20 & 78.63 & 80.54 & 80.21 & 85.51 & 83.63 & 84.36 \\ 
Ariadne's Thread \cite{zhong2023ariadne} & 48.30 & 49.35 & 61.76 & 63.19 & 68.45 & 69.37 & 76.24 & 77.66 & 63.00 & 65.31 & 76.12 & 76.66 \\ \midrule
\rowcolor{gray!15} \multicolumn{13}{c}{\texttt{\textbf{CLIP-Based Approaches}}} \\ 
CLIPSeg \cite{luddecke2022image} & 71.37 & 74.15 & 75.63 & 79.09 & 88.47 & 89.57 & 76.03 & 78.58 & 80.17 & 87.05 & 86.12 & 87.02 \\
DenseCLIP \cite{rao2022denseclip} & 64.62 & 65.78 & 68.83 & 70.21 & 89.17 & 90.18 & 80.16 & 82.29 & 63.93 & 70.74 & 65.81 & 67.52 \\
ZegCLIP \cite{zhou2023zegclip} & 63.76 & 65.79 & 73.54 & 76.35 & 80.58 & 81.47 & 74.98 & 76.83 & 74.90 & 82.17 & 89.47 & 90.21 \\
SAN \cite{xu2023side} & 71.53 & 74.16 & 82.08 & 85.87 & 91.06 & 92.09 & 80.03 & 82.25 & 75.74 & 81.92 & 72.36 & 72.84 \\
MaPLe \cite{khattak2023maple} & 65.58 & 68.06 & 74.13 & 75.66 & 88.51 & 89.51 & 76.56 & 78.71 & 70.15 & 77.35 & 72.68 & 73.42 \\
MaPLe \cite{khattak2023maple} + Decoder & 73.70 & 76.59 & 81.87 & 85.49 & 89.79 & 90.89 & 79.95 & 82.43 & 80.57 & 87.15 & 90.39 & 91.30 \\
VLSM-Adapter \cite{dhakal2024vlsm} & 69.61 & 72.51 & 82.47 & 86.46 & 91.35 & 92.42 & 82.98 & 85.59 & 79.33 & 85.50 & 79.22 & 80.13 \\
CausalCLIPSeg \cite{chen2024causalclipseg} & 68.48 & 70.82 & 75.69 & 79.08 & 88.11 & 89.08 & 73.26 & 75.20 & 76.76 & 83.07 & 85.00 & 85.95 \\
CAT-Seg \cite{cho2024cat} & 72.95 & 75.54 & 83.48 & 85.14 & 90.43 & 91.48 & 83.85 & 85.17 & 80.87 & 87.17 & 88.32 & 89.17 \\   \midrule
\rowcolor{lightpurple}
\texttt{MedCLIPSeg} (Ours) & 81.48 & 84.06 & 85.93 & 89.75 & 91.97 & 93.03 & 88.18 & 90.36 & 82.97 & 89.13 & 92.57 & 93.36 \\
\bottomrule
\end{tabular}
}
\label{tab:full_results_efficiency_50}
\end{table*}

\begin{table*}[t!]
\caption{\textbf{Per-dataset segmentation with 100\% Labeled Data:} 
This table reports DSC and NSD values (\%) across six medical image segmentation benchmarks. 
All baseline methods are trained in a fully supervised manner using ground-truth annotations. Best results are in \textbf{bold}, and second-best are \underline{underlined}.}
\centering
\tablestyle{-27pt}{1.1}
\addtolength{\tabcolsep}{+30pt}
\renewcommand{\arraystretch}{1.5}
\resizebox{0.98\linewidth}{!}{%
\begin{tabular}{ccccccccccccc}
\toprule
\textbf{Method} & \multicolumn{2}{c}{\textbf{BUSI}} & \multicolumn{2}{c}{\textbf{BTMRI}}  & \multicolumn{2}{c}{\textbf{ISIC}} & \multicolumn{2}{c}{\textbf{Kvasir-SEG}} & \multicolumn{2}{c}{\textbf{QaTa-COV19}} & \multicolumn{2}{c}{\textbf{EUS}} \\ 
\cmidrule(lr){2-3} \cmidrule(lr){4-5} \cmidrule(lr){6-7} \cmidrule(lr){8-9} \cmidrule(lr){10-11} \cmidrule(lr){12-13}
& \textbf{DSC} $\uparrow$ & \textbf{NSD} $\uparrow$ & \textbf{DSC} $\uparrow$ & \textbf{NSD} $\uparrow$ & \textbf{DSC} $\uparrow$ & \textbf{NSD} $\uparrow$ & \textbf{DSC} $\uparrow$ & \textbf{NSD} $\uparrow$ & \textbf{DSC} $\uparrow$ & \textbf{NSD} $\uparrow$ 
& \textbf{DSC} $\uparrow$ & \textbf{NSD} $\uparrow$ \\ \midrule
\rowcolor{gray!15} \multicolumn{13}{c}{\texttt{\textbf{Unimodal Approaches}}} \\
UNet \cite{ronneberger2015unet}         & 70.04 & 73.88 & 86.06 & 90.71 & 89.10 & 92.06 & 80.26 & 83.53 & 76.97 & 82.32 & 68.53 & 69.92 \\
UNet++ \cite{zhou2018unet++}       & 67.54 & 71.15 & 83.30 & 87.94 & 89.36 & 92.17 & 84.81 & 88.07 & 73.52 & 78.23 & 72.08 & 73.15 \\
DeepLabv3 \cite{chen2017rethinking}    & 63.02 & 67.18 & 83.49 & 89.31 & 76.34 & 80.24 & 82.00 & 85.73 & 69.63 & 75.67 & 65.21 & 66.37 \\
AttnUNet \cite{oktay2018attention}     & 62.65 & 66.21 & 85.24 & 89.76 & 89.00 & 92.03 & 78.59 & 81.79 & 73.57 & 78.71 & 68.76 & 70.13 \\
nnUNet \cite{isensee2021nnu}       & 76.85 & 80.70 & 86.91 & 92.00 & 90.52 & 93.37 & 85.44 & 89.29 & 75.43 & 81.00 & 73.27 & 74.09 \\
Swin-UNet \cite{cao2022swin}  & 50.37 & 56.13 & 65.34 & 72.51 & 88.10 & 91.00 & 58.69 & 63.87 & 69.69 & 72.43 & 57.96 & 59.95 \\
TransUNet \cite{chen2021transunet} & 57.98 & 62.50 & 70.90 & 76.46 & 88.10 & 91.00 & 58.69 & 63.87 & 68.74 & 71.99 & 58.88 & 61.10 \\ \midrule
\rowcolor{gray!15}
\multicolumn{13}{c}{\texttt{\textbf{Generic Text-driven Approaches}}} \\
% LAVT \cite{yang2022lavt} & 79.26 & 81.87 & 83.38 & 86.83 & 90.44 & 91.47 & 74.55 & 76.58 & 80.32 & 86.39 & 85.07 & 85.87 \\
LViT \cite{li2023lvit} & 75.32 & 77.99 & 81.41 & 84.80 & 91.21 & 92.22 & 85.29 & 87.30 & 82.31 & 87.80 & 84.53 & 85.23 \\ 
Ariadne's Thread \cite{zhong2023ariadne} & 57.26 & 58.22 & 69.96 & 71.40 & 68.37 & 69.30 & 77.42 & 78.79 & 70.70 & 73.94 & 76.71 & 77.31 \\ \midrule
\rowcolor{gray!15} \multicolumn{13}{c}{\texttt{\textbf{CLIP-Based Approaches}}} \\ 
CLIPSeg \cite{luddecke2022image} & 80.95 & 83.87 & 85.33 & 89.45 & 90.55 & 91.62 & 81.98 & 84.61 & 81.76 & 87.30 & 88.66 & 89.58 \\
DenseCLIP \cite{rao2022denseclip} & 71.85 & 74.39 & 70.30 & 72.34 & 89.29 & 90.32 & 79.32 & 81.37 & 65.84 & 72.72 & 68.52 & 70.17 \\ 
ZegCLIP \cite{zhou2023zegclip} & 72.08 & 74.45 & 76.65 & 79.77 & 81.45 & 82.33 & 78.46 & 80.43 & 75.42 & 82.59 & 89.83 & 90.54 \\
SAN \cite{xu2023side} & 77.99 & 80.75 & 85.27 & 89.14 & 91.39 & 92.41 & 83.16 & 85.23 & 76.81 & 82.88 & 75.07 & 75.67 \\
MaPLe \cite{khattak2023maple} & 66.37 & 68.92 & 75.40 & 76.83 & 88.31 & 89.30 & 76.12 & 78.27 & 70.40 & 77.52 & 70.98 & 71.75 \\
MaPLe \cite{khattak2023maple} + Decoder & 80.49 & 83.38 & 85.08 & 89.20 & 90.10 & 91.21 & 83.46 & 85.96 & 81.86 & 88.16 & 88.65 & 89.55 \\
VLSM-Adapter \cite{dhakal2024vlsm} & 80.90 & 83.71 & 85.03 & 89.01 & 91.30 & 92.38 & 85.89 & 88.34 & 81.15 & 87.10 & 78.82 & 79.76 \\
CausalCLIPSeg \cite{chen2024causalclipseg} & 76.11 & 78.70 & 81.71 & 85.30 & 89.47 & 90.46 & 78.77 & 80.79 & 75.67 & 82.37 & 86.30 & 87.59 \\  
CAT-Seg \cite{cho2024cat} & 81.83 & 84.52 & 84.86 & 86.52 & 91.27 & 92.34 & 86.43 & 88.83 & 82.82 & 88.60 & 88.18 & 89.07 \\ \midrule
\rowcolor{lightpurple}
\texttt{MedCLIPSeg} (Ours) & 85.72 & 88.35 & 88.03 & 91.78 & 92.54 & 93.58 & 90.15 & 92.32 & 83.41 & 89.17 & 92.11 & 92.89 \\
\bottomrule
\end{tabular}
}
\label{tab:full_results_efficiency_100}
\end{table*}